\documentclass{llncs}
\usepackage{etex}
\usepackage{setspace}
\usepackage{amsmath} % assumes amsmath package installed
\usepackage{amssymb,nicefrac}
\usepackage[dvips]{epsfig}
\usepackage[dvips]{graphicx}
\usepackage[section]{placeins}
\usepackage{fancyhdr}
\usepackage{color}
\usepackage{pstricks}
\usepackage{pst-node}
\usepackage{dsfont}
\usepackage{mathrsfs}
\usepackage{sgame}
\usepackage{float}
\usepackage[T1]{fontenc}
\usepackage{aurical}
\usepackage{tikz,pgfplots,verbatim}
\usetikzlibrary{spy,arrows,backgrounds,plotmarks}
\usepackage{algorithm}
\usepackage[noend]{algpseudocode}
\usepackage{multirow,bigdelim}

\newif\iffigures
\figurestrue % uncomment if you want figures
\iffigures
\usepgfplotslibrary{external} 
\newcommand{\df}{\doteq}
\newcommand{\magn}[1]{\left\vert #1 \right\vert}

\newcommand{\BLUE}[1]{\textcolor{blue}{#1}}
\newcommand{\VIOLET}[1]{\textcolor{violet}{#1}}
\newcommand{\RED}[1]{\textcolor{red}{#1}}

\begin{document}

\title{An Evolutionary Stochastic-Local-Search Framework for One-Dimensional Cutting-Stock Problems
\thanks{Preliminary versions of part of this paper appeared in \cite{chasparis_optimization_2016}, \cite{rossbory_parallelization_2016}.}
\thanks{The research reported in this article has been supported by the Austrian Ministry for Transport, Innovation and Technology, the Federal Ministry of Science, Research and Economy, and the Province of Upper Austria in the frame of the COMET center SCCH. Partially this work has also been supported by the European Union grant EU H2020-ICT-2014-1 project RePhrase (No. 644235)}
}
% \subtitle{Do you have a subtitle?\\ If so, write it here}

\titlerunning{An Evolutionary SLS Framework for One-Dimensional Cutting-Stock Problems}        % if too long for running head

\author{Georgios C. Chasparis \and Michael Rossbory \and Verena Haunschmid %etc.
}

%\authorrunning{Short form of author list} % if too long for running head

\institute{   Software Competence Center Hagenberg GmbH \\
              Softwarepark 21 \\
              4232 Hagenberg, Austria \\
              Tel.: +43 7236 3343 857\\
              Fax: +43 7236 3343 888\\
              \email{\{georgios.chasparis,michael.rossbory,verena.haunschmid\}@scch.at}           %  \\
%             \emph{Present address:} of F. Author  %  if needed
%           \and
%           S. Author \at
%              second address
}

\date{Received: \today / Accepted: date}
% The correct dates will be entered by the editor

\maketitle

\begin{abstract}
We introduce an evolutionary stochastic-local-search (SLS) algorithm for addressing a generalized version of the so-called 1/V/D/R cutting-stock problem. Cutting-stock problems are encountered often in industrial environments and the ability to address them efficiently usually results in large economic benefits. Traditionally linear-programming-based techniques have been utilized to address such problems, however their flexibility might be limited when nonlinear constraints and objective functions are introduced. To this end, this paper proposes an evolutionary SLS algorithm for addressing one-dimensional cutting-stock problems. The contribution lies in the introduction of a flexible structural framework of the optimization that may accommodate a large family of diversification strategies including a novel parallel pattern appropriate for SLS algorithms (not necessarily restricted to cutting-stock problems). We finally demonstrate through experiments in a real-world manufacturing problem the benefit in cost reduction of the considered diversification strategies.
% \PACS{PACS code1 \and PACS code2 \and more}
% \subclass{MSC code1 \and MSC code2 \and more}
\end{abstract}


%%%%%%%%%%%%%%%%%%%%%%%%%%%%%%%%%%%%%%%%%%%%%%%%%%%%%%%%%%%%%%%%%%%%%%%%%%%%%%%%%%%%%%%%%%%%%%%%%
\section{Introduction} \label{sec:Introduction}

Cutting-stock problems formulate economic optimization problems aiming at the minimization of stock use so that certain job requirements are satisfied. They are concerned with the efficient slitting of bands of material out of a set of given rolls of material. Apparently, due to the immediate economic incentives, cutting-stock problems have attracted considerable attention, especially with respect to the development of optimization techniques for more efficient solutions. Cutting-stock problems have a long history, starting with the first known formulation of \cite{Kantorovich60}, and the first advanced solution procedures based on linear programming proposed in \cite{Gilmore61,Gilmore63}. % \cite{Dyckhoff90} developed a typology for classifying cutting-stock and packing problems.

In industrial environments, such as in the electrical transformers industry \cite{chasparis_optimization_2016}, linear-programming techniques may not be appropriate due to the large number of constraints, some of which may be nonlinear in the parameters. Examples of such nonlinear constraints may include constraints imposed through eco-design measures (cf.,~\cite{chasparis_optimization_2016}). Due to these nonlinear constraints, heuristic-based optimization techniques seem more appropriate for addressing this family of problems. To this end, this paper introduces a stochastic-local-search (SLS) algorithm for addressing a class of one-dimensional cutting-stock problems, namely the category 1/V/D/R according to the typology of \cite{Dyckhoff90}.

In the remainder of this section, Section~\ref{sec:RelatedWork} presents related work and Section~\ref{sec:Contribution} states the main contribution of this paper. In the remainder of this paper, Section~\ref{sec:ProblemFormulation} presents the class of the cutting-stock problems addressed by this paper and Section~\ref{sec:EvolutionaryStochasticLocalSearchAlgorithm} presents the proposed evolutionary SLS algorithm. In Sections~\ref{sec:Operations}--\ref{sec:DiversificationStrategies}, we provide a more detailed description of the building blocks of the proposed algorithm, namely the set of possible operations for local search in Section~\ref{sec:Operations} and the set of diversification strategies in Section~\ref{sec:DiversificationStrategies}. Finally, in Section~\ref{sec:ExperimentalEvaluation}, we provide an experimental evaluation of the proposed algorithm, and Section~\ref{sec:Conclusions} presents concluding remarks.

\subsection{Related work} \label{sec:RelatedWork}

The one- or two-dimensional cutting stock problem is the problem of optimizing the slitting of material into smaller pieces of predefined sizes subject to several constraints. Almost 80 years ago the cutting stock problem was first addressed by Kantorovich in 1939 (translated to English in 1960~\cite{Kantorovich60}) and Brooks in 1940 (reprinted in 1987~\cite{Brooks87}) (but not yet using that name). This problem first arose in the field of cutting paper rolls~\cite{Sweeney90} and was later also used for the processing of metal sheets~\cite{Gerstl97}, wood~\cite{Cherri14}, glass, plastics, textiles and leather~\cite{Dyckhoff90}. More recent publications also deal with stent manufacturing~\cite{Aktin09} and cutting liquid crystal display panels from a glass substrate sheet~\cite{Lu15}.

The cutting stock problem is closely related to bin packing, strip packing, knapsack, vehicle loading and several other problems~\cite{Dyckhoff90}. A very commonly used objective is trim-loss or waste minimization. Other potential objectives that are mentioned in~\cite{Aktin09} can be the minimization of a generalized total cost function (consisting of material inputs, number of setups, labour hours and overdue time), subject to material availability, overtime availability and date constraints. Another interesting characteristic is the possibility to have multiple stock lengths~\cite{Belov02,Holthaus02} or the possibility of using leftovers~\cite{Cherri14}. An additional difficulty is when both the master rolls and the customer order can have multiple quality gradations~\cite{Sweeney90}. 

%The similarity is the logical structure of cutting stock with packing bins and the other problems mentioned. 

% typology

Due to the fact that many different types of cutting stock problems can be found in the literature, e.g. with respect to dimensionality, characteristics of large and small objects, shape of figures and so forth, it is highly desirable that the scientific community is using the same language and terminology when describing problems and the corresponding solution approaches. In \cite{Dyckhoff90} a typology that covered the most important four characteristics at the time to classify optimization problems like the cutting stock problems was introduced. This typology was improved by~\cite{Wascher07}. An overview of characteristics how to classify the problems in the literature can be found in \cite{Pentico08}.
The problem can further be distinguished whether it is off-line (full knowledge of the input) or on-line (no knowledge of the next items)~\cite{Lodi02}. 

% solving them

The cutting stock problem can be expressed as an integer programming problem with the downside that the large number of variables involved makes the computation of an optimal solution infeasible~\cite{Gilmore61}. To solve such integer optimization problems a broad range of algorithms exist. First advanced solution procedures based on linear programming were proposed by \cite{Gilmore61,Gilmore63}. 
A review and classification of approaches into heuristics, item-oriented or cutting pattern-oriented can be found in~\cite{Cherri14}.

Most publications in this field employ exact algorithms which guarantee to find the optimal solution~\cite{Hoos05}. Exact solution approaches comprise column generation~\cite{Holthaus02,Kallrath14}, branch-and-bound methods~\cite{Gerstl97} or a combination of them, e.g., the so-called branch-and-price algorithm~\cite{Belov02,Belov06}. A review of several linear programming formulations for the one-dimensional cutting stock and bin packing problem can be found in~\cite{Carvalho02}.

Another group of publications is dealing with heuristic solution approaches. Contrary to exact algorithms, heuristic approaches do not guarantee to find the optimal solution~\cite{Hoos05}. Heuristic methods can be split into three categories: sequential heuristic procedures, linear programming based methods and metaheuristics~\cite{Beraldi09}. A procedure that is often mentioned in the literature is the sequential heuristic procedure~\cite{Gradisar99,Sweeney90}. Many other publications deal with heuristics as well.
In~\cite{Gerstl97} a very simple greedy heuristic is used. Some authors use heuristic-based column generation~\cite{Umetani03,Aktin09} or reduction~\cite{Yanasse06} techniques. An example for an approach which combines an exact (branch-and-bound) with a heuristic (sequential heuristic procedure) method can also be found in the literature~\cite{Gradisar05}.

It appears that the first mention of using genetic algorithms for the cutting stock problem can be found in~\cite{Hinterding93}. The authors state that the earliest use of heuristic search techniques in the field of operations research was published in 1985. Whereas reference \cite{Hinterding93} uses an improved version of the traditional chromosome representation, reference \cite{Peter96} represents cutting patterns as tree graphs. Genetic algorithms have also been used for related combinatorical problems, e.g. integrating processing planning and scheduling (usually those problems are considered separately)~\cite{Shao09} and solving the variable sized bin-packing problem~\cite{Haouari09}.
Further genetic algorithm approaches have been published in~\cite{Onwubolu03,Romanowski12}. An overview over previous genetic algorithm applications can be found in~\cite{Shahin04}. The authors also provide a genetic algorithm model which was used to study three real-world case studies. Recently, an efficient genetic algorithm for the cutting stock problem is proposed by \cite{Lu15}. While most publications describe the crossover and mutation operators implemented~\cite{Haouari09,Lu15,Shahin04}, none of them elaborate on diversification strategies which are an essential part of genetic algorithms.

\subsection{Contribution}	\label{sec:Contribution}

The main objective of this paper is the development of an evolutionary SLS algorithm to efficiently address 1/V/D/R cutting-stock problems including nonlinear constraints. In particular, our goal is to provide an optimization framework that a) exploits the capabilities of modern computing systems through parallelization, and b) provides a flexible integration and tuning of a large family of diversification strategies. 

In fact, the proposed optimization framework led to the development of a novel high-level parallel pattern specifically tailored for evolutionary SLS algorithms, an earlier version of which presented by the authors in \cite{rossbory_parallelization_2016}. The proposed diversification strategies are easily incorporated to the structural characteristics of the parallel pattern and demonstrate the flexibility of this framework to accommodate a great variety of settings depending on both user demands and computational capabilities. Finally, we present an experimental evaluation demonstrating the relative cost reduction incurred due to the introduction of the proposed diversification strategies.

A preliminary version of the proposed methodology first introduced by the authors in \cite{chasparis_optimization_2016}, however, without providing the details of the optimization algorithm. The present paper provides a ready-to-use comprehensive treatment of one-dimensional cutting-stock problems usually encountered in the industry.

\subsection{Notation}

For convenience, we summarize here some generic notation that is used throughout the paper.

\begin{itemize}
\item $\mathbb{I}_{A}$ denotes the index function such that 
\begin{equation*}
\mathbb{I}_{A} \doteq \begin{cases}
1 & \mbox{if } A=\mathtt{true}, \cr
0 & \mbox{else.}
\end{cases}
\end{equation*}
\item $e_i\in\mathbb{R}^{n}$ denotes the unit vector of size $n$, such that its $i$th entry is equal to $1$ and the rest are equal to $0$;
\item For a matrix $X=[x_{ij}]_{i\in\mathcal{I},j\in\mathcal{J}}$, we denote the row vector corresponding to the $i$th row as $X_{i:} = [x_{ij}]_{j\in\mathcal{J}}$ and the column vector corresponding to the $j$th column as $X_{:j} = [x_{ij}]_{i\in\mathcal{I}}$.
\item For a finite set $A$, $\magn{A}$ denotes its cardinality.
%\item $\Pi$ denotes the set of instances of the optimization;
%\item $\pi$ denotes a representative element of the set $\Pi$;
%\item $X$ denotes a candidate solution of the optimization;
%\item $V(X)$ denotes the objective function of the optimization.
%\item $\mathcal{I}=\{1,2,...,m\}$ set of available items, with $i$ being a representative element;
%\item $\mathcal{J}=\{1,2,...,n\}$ set of requested objects, with $j$ being a representative element;
%\item $b$ denotes width;
%\item $\ell$ denotes length;
%\item $d$ denotes specific weight (of an object or item);
%\item $w$ denotes weight (of an object or item);
%\item $W_i$ denotes the produced weight of item $i$;
%\item $r_j$ denotes the rest band of object $j$;
% \item $\mathcal{R}_j$ denotes the set of allowable rest bands of object $j$
%\item $N_{\rm pr}$ number of operations in the perturbation worker;
%\item $N_{\rm lo}$ number of operations in the local optimization worker;
%\item $N_{\rm cn}$ number of operations in a constraint worker;
\end{itemize}

%%%%%%%%%%%%%%%%%%%%%%%%%%%%%%%%%%%%%%%%%%%%%%%%%%%%%%%%%%%%%%%%%%%%%%%%%%%%%%%%%%%%%%%%%%%%%%%%%
\section{Problem Formulation}		\label{sec:ProblemFormulation}

We consider a general class of one-dimensional (1/V/D/R) cutting stock problems. In particular, we are given a set of \emph{items} (or \emph{bands}) of certain types $\mathcal{I}=\{1,2,...,n\}$, each of which is characterized by its width $b_i$ and its desired weight $w_i$. The set of desired items $\mathcal{I}$ need to be placed to a set of \emph{objects} (or \emph{rolls}) of material, denoted by $\mathcal{J}=\{1,2,...,m\}$ each of which is characterized by its width $b_j$, its length $\ell_j$ and its specific weight $d_j$, $j\in\mathcal{J}$. Let also $w_j$ denote the overall weight of roll $j$.

% there exists a set of available \emph{objects} (or \emph{rolls}) of material, denoted by $\mathcal{I}\doteq\{1,2,...,m\}$ each of which is characterized by its width $b_i$, its length $\ell_i$ and its specific weight $d_i$, $i\in\mathcal{I}$. Let also $w_i$ denote the overall weight of roll $i$. We are also provided with a set of \emph{items} (or \emph{bands}) of certain types $\mathcal{J}\doteq\{1,2,...,n\}$, each of which is characterized by its width $b_j$ and its desired weight $w_j$. 

\begin{figure}[t!]
\centering
\includegraphics[scale=1]{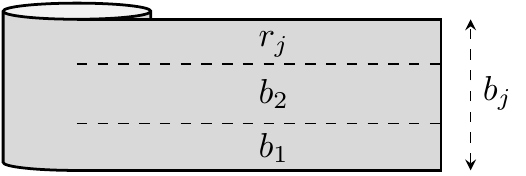}
\caption{Roll slitting.}
\label{fig:rollslitting}
\end{figure}

We define an assignment of bands into rolls, denoted $x_{ij}\in\mathbb{Z}_+$, so that $x_{ij}$ denotes the number of bands of type $i$ assigned to roll $j$. The objective is to find an assignment $X\doteq\{x_{ij}\}_{i,j}$ of bands into the available rolls, so that:
\begin{enumerate}
\item the overall weight of each band type $i$ exceeds its desired weight $w_i$, i.e., $$b_i\sum_{j=1}^{m}x_{ij}\ell_j d_j \geq w_i.$$ To simplify notation, let us denote $\alpha_{j} \doteq \ell_jd_j$, in which case the constraint above may equivalently be written as
\begin{equation}
y_i(X) \df w_i - b_i\sum_{j=1}^{m}x_{ij}\alpha_{j} \leq 0,
\end{equation}
where $y_i(X)$ corresponds to the $\emph{rest weight}$ of item $i$.
% In several cases, we will use the notation $W^*\df \sum_{j=1}^{n}w_j$ to denote the overall desired \emph{job weight} over all bands.
\item the overall width assigned to each roll $i$ does not exceed the width of the roll, $b_i$, while at the same time the residual band, denoted 
\begin{equation}
r_j(X) \doteq b_j - \sum_{i=1}^{n}x_{ij}b_i \in \mathcal{R}_j,
\end{equation}
where $\mathcal{R}_j\subseteq[0,\infty)$ is a union of closed intervals of allowable residual bands, which might be different for each roll $j\in\mathcal{J}$.
\end{enumerate}

Both of the above constraints are hard constraints and need to be satisfied for any assignment. Unfortunately, in most but trivial cases, there might be a multiplicity of admissible assignments, each of which might be utilizing a different subset of rolls $\mathcal{J}$. We wish to minimize the overall weight of the rolls utilized by an assignment, i.e., we wish to address the following optimization problem:
\begin{equation*}	
\min_{X\in\mathbb{Z}_{+}^{m\times{n}}} g(X),
\end{equation*}
where
\begin{equation}	\label{eq:ObjectiveFunction}
g(X) \df \sum_{j\in\mathcal{J}}w_j\mathbb{I}_{\{\exists i\in\mathcal{I}:x_{ij}>0\}}
\end{equation}
is the total weight of rolls used by assignment $X$. %, where $\mathbb{I}_{A}$ denotes the index function such that 
%\begin{equation*}
%\mathbb{I}_{A} \doteq \begin{cases}
%1 & \mbox{if } A=\mathtt{true}. \cr
%0 & \mbox{else.}
%\end{cases}
%\end{equation*} 
Then, the overall optimization problem may be written in the following form:
\begin{subequations} \label{eq:Optimization}
\begin{align}
\min_{X} & \quad g(X) && \mbox{(\emph{objective})} \label{eq:Objective} \\
\mbox{s.t.} & \quad  y_i(X) \leq 0 && \mbox{(\emph{job-admissibility constraint}, $\mathcal{C}_{\rm job}$)} \label{eq:JobAdmissibility} \\
& \quad  r_j(X) \in \mathcal{R}_j && \mbox{(\emph{rest-width constraint}, $\mathcal{C}_{\rm rw}$)} \label{eq:RestWidthAdmissibility} \\
\mbox{var.} & \quad  X\in\mathbb{Z}_{+}^{n\times{m}} && \mbox{(\emph{domain})} \label{eq:OptimizationAssignment}
\end{align}
\end{subequations}

Note that we allow for $\mathcal{R}_j$ to depend on the roll $j\in\mathcal{J}$. This is due to the fact that some extra processing of the rolls may be necessary for each item which depends on the roll itself (e.g., seaming of the edges). Furthermore, we do not allow all possible residual bands in $\mathcal{R}_j$, since not all residual bands can be stored in stock. This constraint introduces a nonlinearity. 

The details of the optimization problem (\ref{eq:Optimization}) (i.e., the characteristics of items $\mathcal{I}$ and objects $\mathcal{J}$ and the set of allowable rest-widths $\mathcal{R}_j$, $j\in\mathcal{J}$), define an instance of the optimization problem, denoted $\pi$. Let also $\Pi$ be the family of such optimization instances.

We have currently chosen a rather small number of constraints, namely the job-admissibility and rest-width constraint. However, it is important to point out that additional constraints may also be added, depending on the application of interest, such as the eco-design measures mentioned in \cite{chasparis_optimization_2016}. Such constraints may be nonlinear in the parameters. Furthermore, alternative objective criteria may be considered (e.g., minimization of trim loss, as considered in \cite{Haessler91}, or minimization of rolls needed by the assignment when the objects are identical, as considered in \cite{Carvalho02}). 

It is important to note that the forthcoming algorithm does \emph{not} depend on the specifics of the considered constraints in optimization (\ref{eq:Optimization}). In fact, there is no restriction on the type and number of constraints imposed. Thus, the reader should consider optimization problem (\ref{eq:Optimization}) more as an example that will convenience the discussion throughout the paper, rather than as a restriction.

\section{Evolutionary Stochastic-Local-Search (SLS) Algorithm}	\label{sec:EvolutionaryStochasticLocalSearchAlgorithm}

The proposed algorithm consists of the following phases:
\begin{itemize}
\item Initialization (\texttt{init});
\item Optimization (\texttt{optimize});
\item Filtering (\texttt{filter});
\item Selection (\texttt{select}).
\end{itemize}

The role of the \emph{initialization} (\texttt{init}) phase is the establishment of candidate solutions (not necessarily satisfying all the constraints). The performance of the initialization function is rather important to the performance of the overall optimization, since a good head-start may significantly improve the performance of the SLS algorithm. 

The role of the \emph{optimization} (\texttt{optimize}) phase is the execution of admissible local improvement steps, accompanied with admissible perturbations. Responsible for the execution of these improvement steps and/or perturbations of the existing candidate solutions are the \emph{working units}, namely \texttt{workers}, each of which plays a distinctive role (explained in detail in the forthcoming Section~\ref{sec:Optimize}). We may argue that the core of the proposed architecture lies in the design of the working units.

The role of the \emph{filtering} (\texttt{filter}) phase is the observation of the current status of the optimization algorithm, the storage and/or re-processing of candidate solutions. In other words, it supervises the state of all candidate solutions and decides on their future use. 

Finally, the role of the \emph{selection} (\texttt{select}) phase is to decide on whether the optimization algorithm should be terminated according to some criteria (e.g., the size of the pool or time constraints) and, if yes, what should be the best estimate $\hat{X}$ of the optimal solution. 

\begin{figure}[t!]
\centering
\includegraphics[scale=0.65]{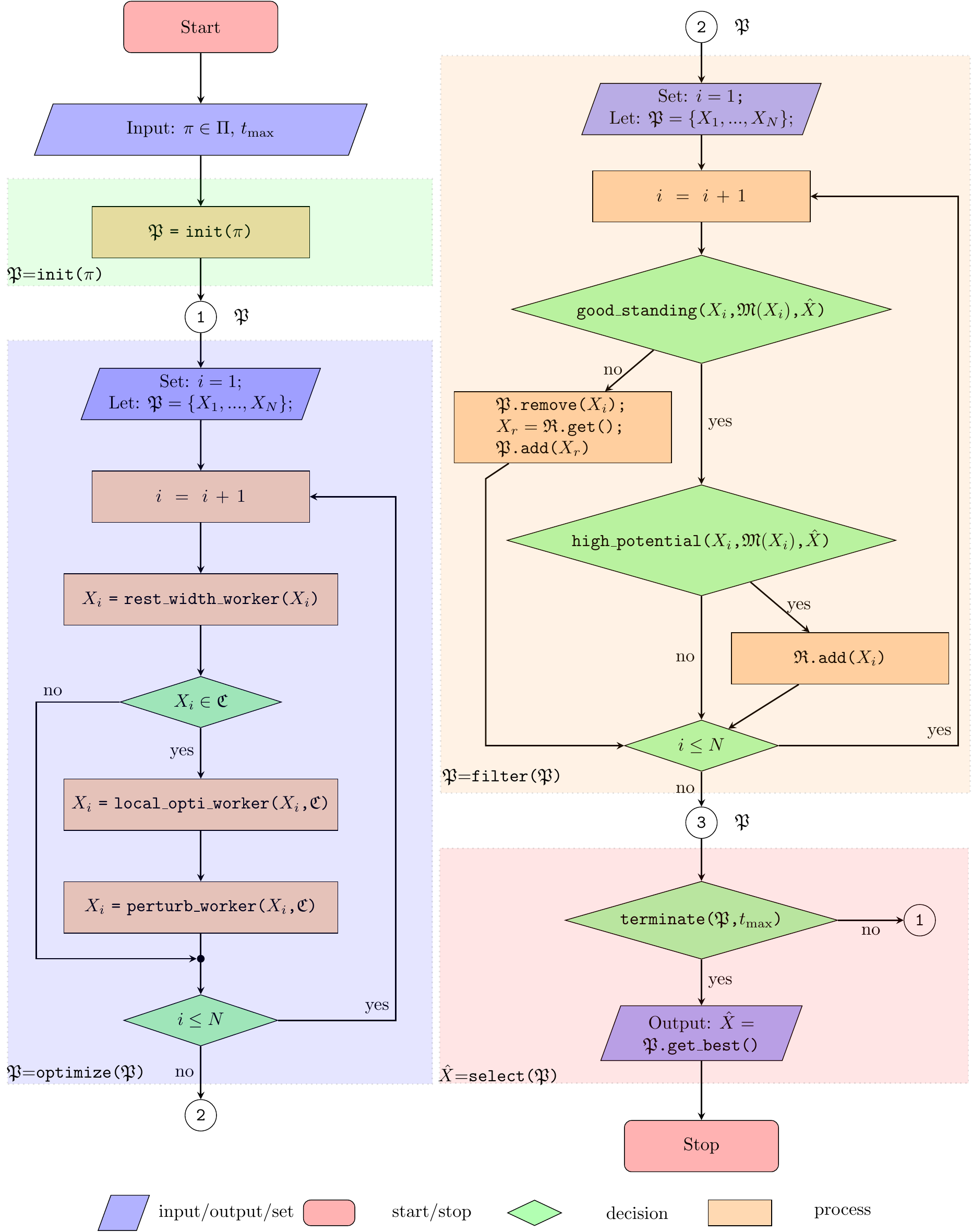}
\caption{Architecture of optimization algorithm.}
\label{fig:architecture}
\end{figure}

The details of the (SLS) optimization algorithm are presented in Figure~\ref{fig:architecture}, and they will be explained progressively in the forthcoming sections. % Furthermore, we have gathered the main notation used by the optimization algorithm in Table~\ref{Tb:Notation}, explained in detail in the remainder of this section.

\subsection{Initialization (\texttt{init})}	\label{sec:Initialization}

The initialization phase discovers initial \emph{candidate solutions} (i.e., assignments $X$ which satisfy the job-admissibility constraint). The benefit of discovering such candidate solutions is rather important since it may fasten significantly the following optimization phase. Furthermore, if there exist a large number of \emph{local optima}\footnote{We use the term \emph{local optima} to refer to assignments $X$ at which any local search (based on some predefined operations/modifications) may not result in a \emph{strictly better} assignment with respect to the objective function.} (as it is often the case), the starting point may influence significantly the resulting performance. To this end, it is necessary that we consider alternative fitness functions for the generation of candidate solutions.

The goal of the initialization routine is to create an initial set of job-admissible assignments $X$, denoted $\mathfrak{P}$ (\emph{main pool}) of bounded maximum size ($\magn{\mathfrak{P}}_{\max}$). This pool of candidate solutions will then be used for optimization.  The possibilities for designing such a routine are, in fact, open ended. Here, we would like to provide one general structure, the specifics of which are provided in Table~\ref{Tb:Initialization}. 

\begin{table}[t!]
\boxed{\footnotesize{
\begin{minipage}{0.98\textwidth}
\begin{tabular}{ll}
\hline
\textbf{function} \texttt{$\mathfrak{P}$ = init($\pi$)} & \\\hline
%\texttt{//check if an item fits exactly into an object} & \\
%%\texttt{\BLUE{SET} $k$ as the maximum number of items in an object} & \\
%\texttt{$X$ = zeros($m\times{n}$)} & \mbox{(initializing assignment)} \\
%\texttt{\BLUE{FOR} ($j\in\mathcal{J}$) } & \\
%\quad \texttt{\BLUE{FOR} ($i\in\mathcal{I}$) } & \\
%\quad \quad \texttt{\BLUE{IF} ($b_i$ = $b_j$) } \\
%\quad \quad \quad $X_{:j}=X_{:j}+e_i;$ & \mbox{(assign item $i$ into $j$)} \\
%\quad \quad \texttt{\BLUE{END IF}} & \\
%\quad \texttt{\BLUE{END FOR}} & \\
%\texttt{\BLUE{END FOR}} & \\\hline
\texttt{//assignment based on criterion of (\ref{eq:InitializationFitness1}),(\ref{eq:InitializationFitness2}) or (\ref{eq:InitializationFitness3})} & \\
\texttt{\BLUE{FOR} (every item $i$ in $\mathcal{I}$ (descending order)) } & \\
\quad \texttt{\BLUE{SET} $f_i^* = -\infty$ } & \\
\quad \texttt{\BLUE{FOR} (every object $j$ in $\mathcal{J}$ (descending order)) } & \\
%\quad \quad \texttt{\BLUE{FOR} (each combination $c\in\mathcal{C}_k$) } & \\
%\quad \quad \texttt{\BLUE{SET} applies=\BLUE{TRUE}} & \\
%\quad \quad \texttt{\BLUE{FOR} (each item $i$ in combination $c$) } & \\
%\quad \quad \quad \texttt{\BLUE{IF} ($y_i\leq{0}$) } & \\
%\quad \quad \quad \quad \texttt{applies=\BLUE{FALSE};} & \\
%\quad \quad \quad \quad \texttt{\BLUE{BREAK};} & \\
%\quad \quad \quad \texttt{\BLUE{END IF}} & \\
\quad \quad \texttt{\BLUE{IF} ($y_i(X)>0$ and $r_j(X)>b_i$) } & \\
\quad \quad \quad $X_{:j} = X_{:j} + e_i$; & \mbox{(assign item $i$ to object $j$)} \\
\quad \quad \quad \texttt{\BLUE{IF} ($f_i(X;j) > f_i^*$) } & \\
\quad \quad \quad \quad $f_i^* = f_i(X;j)$;  & \mbox{(accept item $i$)} \\
\quad \quad \quad \texttt{\BLUE{ELSE}} & \\
\quad \quad \quad \quad $X_{:j} = X_{:j} - e_i$; & \mbox{(remove item $i$)} \\
\quad \quad \quad \texttt{\BLUE{END IF}} & \\
\quad \quad \texttt{\BLUE{END IF}} & \\
\quad \texttt{\BLUE{END FOR}} & \\
\texttt{\BLUE{END FOR}} & \\
\texttt{//populate main pool of candidate solutions} & \\
\texttt{\BLUE{SET} $\mathfrak{P}$ = (multiple copies of) $X$} & \\
\end{tabular}
\end{minipage}
}
}
\caption{\texttt{init}}
\label{Tb:Initialization}
\end{table}

% We first set $k\in\mathbb{N}$ to be the maximum number of items that an object may fit (which is usually known given the nature of the problem). We then generate the set of combinations (with repetitions) of $k$ items out of the set $\mathcal{I}$, denoted $\mathcal{C}_k$. Finally, given a current assignment $X$, we check which combination of items, $c\in\mathcal{C}_k$, fits ``\emph{better}'' into an object $j\in\mathcal{J}$. The criterion based on which the best fit combination is decided is determined by an fitness criterion denoted $f_j(c;X)$, which measures the fitness of combination $c$ into object $j$ under assignment $X$. 

As a first step, we sort both items $\mathcal{I}$ and objects $\mathcal{J}$ in descending order with respect to their width, similarly to a standard First-Fit-Decreasing (FFD) algorithm \cite{williamson_design_2011}. This is usually necessary, at least in cutting-stock problems, since large items may only fit in large objects. Then, for each item $i$ (with positive rest weight), we go through each available object $j$ (which may fit $i$), and we check whether an assignment of $i$ to $j$ maximizes a fitness criterion. In particular, we introduce here three candidate \emph{fitness} functions:
\begin{enumerate}
\item negative weight of residual band of object $j$, i.e., 
\begin{equation}	\label{eq:InitializationFitness1}
f_i(X;j) \doteq - r_j(X)\alpha_j.
\end{equation}
This fitness function is maximized when the weight of the residual band takes values that are close to zero. In other words, we try to fit as many objects as possible in each object $j\in\mathcal{J}$.
\item negative absolute rest weight of item $i$ minus the weight of the residual band of object $j\in\mathcal{J}$, i.e., 
\begin{equation}	\label{eq:InitializationFitness2}
f_i(X;j) \doteq - |y_i(X)| - r_j(X)\alpha_j.
\end{equation}
Note that this function takes large values when both a) the rest weight of item $i$ and b) the residual weight of object $j$ are close to zero. In other words, under this criterion, we try to fit as many items as possible within object $j$ without though generating large rest weight for the selected items.
\item band $i$'s weight minus the weight of object $j$'s residual band, i.e.,
\begin{equation}	\label{eq:InitializationFitness3}
f_i(X;j) \doteq \alpha_jb_i - \alpha_jr_j(X).
\end{equation}
Note that this function takes large values when the weight of item $i$ is large compared to the residual band's weight, i.e., we try to fit as many bands as possible in each single object $j$.
\end{enumerate}

Alternative fitness functions may be defined. Note that all the above fitness criteria employ a form of weight optimization (contrary to standard First-Fit-Decreasing algorithms). Such form of weight optimization is motivated primarily by two practical reasons: a) the difficulty in finding admissible assignments when the stock material is limited (something that cannot be guaranteed by standard First-Fit-Decreasing algorithms), and b) the limited time offered to solve the optimization problem, which makes the initial assignment a rather important factor over the final performance. In the forthcoming experimental evaluation (Section~\ref{sec:ExperimentalEvaluation}), we will demonstrate the impact of this good head-start in the final performance.

In the remainder of this paper, we will assume that an initial candidate solution $X$ that satisfies the job-admissibility constraint is available as a result of this initialization phase. Note that we do not impose the rest-width constraint during this initialization phase, due to the fact that in several cases it is rather difficult to meet this constraint. A special treatment of the rest-width constraint is suggested as part of the following optimization phase.

\subsection{Optimization (\texttt{optimize})}	\label{sec:Optimize}

As described in Figure~\ref{fig:architecture}, the results of the \texttt{init} routine are used to populate an initial set of candidate solutions (\emph{main pool} $\mathfrak{P}$) that will later be processed with \texttt{optimize} to generate admissible solutions for the optimization instance $\pi\in\Pi$.

In particular, the following steps are executed. First, the available candidate solutions in $\mathfrak{P}$ are sequentially processed through a sequence of \emph{working units}. They are responsible for performing the necessary steps for either establishing admissible solutions and/or for further improving the performance of a candidate solution. We introduce three types of working units, namely:
\begin{itemize}
\item \texttt{rest\_width\_worker($X$,$\mathfrak{C}$)}
\item \texttt{local\_opti\_worker($X$,$\mathfrak{C}$)}
\item \texttt{perturb\_worker($X$,$\mathfrak{C}$)}
\end{itemize}
where $\mathfrak{C}\df\{\mathcal{C}_{\rm job},\mathcal{C}_{\rm rw}\}$ is the family of constraints considered, corresponding to the job-admissibility (\ref{eq:JobAdmissibility}) and rest-width (\ref{eq:RestWidthAdmissibility}) constraints. 

The role of the \texttt{rest\_width\_worker($X$,$\mathfrak{C}$)} is to address the rest-width admissibility constraint, since it might not easily be satisfied for  the job-admissible assignments generated at \texttt{init}. In case additional constraints were considered in the optimization problem (\ref{eq:Optimization}), then additional such workers would have to be introduced.

The role of the \texttt{local\_opt\_worker($X$,$\mathfrak{C}$)} is to generate improved assignments $X$ with respect to the objective function in (\ref{eq:Optimization}).

Finally, the role of the \texttt{perturb\_worker($X$,$\mathfrak{C}$)} is to perform a sequence of perturbations in the currently admissible solutions in order to avoid convergence to local optima.

All these classes of working/processing units are based upon appropriately introduced modifications of the candidate solution $X$, so that either a constraint is satisfied or an improvement is observed in the cost function. These modifications, namely \emph{operations}, constitute the core of the proposed optimization scheme and will be described in detail in the forthcoming Section~\ref{sec:Operations}. A set of available operations will be considered, denoted $\mathcal{O}$. Let also $o\in\mathcal{O}$ denote a representative element of this set. For now, we will only assume that each operation $o\in\mathcal{O}$ is equipped with three alternative modes, denoted:
\begin{itemize}
\item \texttt{better\_reply($X$,$\mathfrak{C}$,$g()$)},
\item \texttt{constr\_reply($X$,$\mathfrak{C}$,$\mathcal{B}$)},
\item \texttt{random\_reply($X$,$\mathfrak{C}$)}.
\end{itemize}
The above introduced modes designate three alternative possibilities for searching and executing operations in the candidate solution $X$. The better reply of an operation $o\in\mathcal{O}$, denoted  \texttt{better\_reply($X$,$\mathfrak{C}$,$g()$)}, is based upon a semi-stochastic search of neighboring assignments to $X$ which improves the cost of the current assignment $X$, i.e., it generates an improved candidate solution with respect to the objective $g()$ that also satisfies the constraints in $\mathfrak{C}$. The \texttt{constr\_reply($X$,$\mathfrak{C}$,$\mathcal{B}$)} mode performs a similar type of search, however the objective is to guarantee that the resulting assignment satisfies the constraints within $\mathfrak{C}$ of the optimization. The set $\mathcal{B}\subset\mathcal{J}$ is the set of ``bad'' objects that currently do not satisfy the constraint. Such a mode may be particularly beneficial when $X$ does not currently satisfy some of the constraints. Finally, the \texttt{random\_reply($X$,$\mathfrak{C}$)} mode of an operation $o\in\mathcal{O}$ is based upon a \emph{random} generation of local to $X$ assignments that satisfy the constraints $\mathfrak{C}$, however they may not necessarily improve the cost of $X$.

In the remainder of this section, we describe in detail the three introduced working units of the optimization routine which utilize some or all of the three alternative operation modes stated above.

\subsubsection*{Rest-width worker (\texttt{rest\_width\_worker})}	\label{RestWidthWorker}

Note that the initially generated assignments $X$ in $\mathfrak{P}$ define candidate solutions that satisfy only the job-admissibility constraint (\ref{eq:JobAdmissibility}). The role of the rest-width worker is to operate on the candidate solutions on the main pool $\mathfrak{P}$, so that they satisfy the rest-width admissibility constraint for all objects $j\in\mathcal{J}$, without violating the job-admissibility constraint. The rest-width constraint is usually the most difficult constraint to be satisfied, and this is the reason it cannot be treated during the initialization phase of Section~\ref{sec:Initialization}. Usually, a sequence of operations might be necessary before rest-width admissibility is achieved. 

The rest-width worker is governed by the parameter $N_{\rm con}$ which corresponds to the total number of processing steps executed per each run of this worker. The architecture of the rest-width worker is presented in Table~\ref{Tb:RestWidthWorker}. 
\begin{table}[h!]
\boxed{\footnotesize{
\begin{minipage}{0.98\textwidth}
\begin{tabular}{ll}
\hline
\textbf{function} \texttt{$X$ = rest\_width\_worker($X$,$\mathfrak{C}$)} & \\\hline
\texttt{\BLUE{REPEAT} (for $N_{\rm con}$ times) } & \\
\quad \texttt{\BLUE{SET} $\mathcal{B}=\{j\in\mathcal{J} \mbox{ such that } r_j(X)\notin\mathcal{R}_j \}$} & (find ``bad'' objects) \\
\quad \texttt{\BLUE{IF} ($\mathcal{B} = \varnothing$)} & (if there are no ``bad'' objects)\\ 
\quad \quad \texttt{\BLUE{RETURN} $X$} & (return current assignment) \\
\quad \texttt{\BLUE{ELSE} } & (if there are ``bad'' objects)\\
\quad \quad \texttt{\BLUE{FOR} (\RED{oper}=0 to \RED{oper}<$\magn{\mathcal{O}}$)} & (for each operation in $\mathcal{O}$)\\
\quad \quad \quad \texttt{\BLUE{SET} $\mathcal{B}=\{j\in\mathcal{J} \mbox{ such that } r_j(X)\notin\mathcal{R}_j \}$} & (find ``bad'' objects)\\
\quad \quad \quad \texttt{\BLUE{IF} ($\mathcal{B}=\varnothing$)} & \\ 
\quad \quad \quad \quad \texttt{\BLUE{RETURN} $X$} & \\
\quad \quad \quad \texttt{\BLUE{ELSE}} & \\
\quad \quad \quad \quad \texttt{$X=\mathcal{O}$[\RED{oper}].constr\_reply($X$,$\mathfrak{C}$,$\mathcal{B}$)} & (execute \texttt{constr\_reply} mode)\\
\quad \quad \quad \texttt{\BLUE{END IF}} & \\
\quad \quad \texttt{\BLUE{END FOR}} & \\
\quad \texttt{\BLUE{END IF}} & \\
\texttt{\BLUE{END REPEAT}} & \\
\end{tabular}
\end{minipage}
}
}
\caption{\texttt{rest\_width\_worker}}
\label{Tb:RestWidthWorker}
\end{table}
As we can see, it simply executes a number of operations under the \texttt{constr\_reply} mode until a rest-width admissible solution is found.

It is important to point out that if a candidate solution $X$ satisfies this constraint (after passing through this worker), then it will satisfy this constraint for the remaining of its processing history, i.e., no further processing will be requested for $X$ within this working unit. This is due to the fact that any subsequent working unit will never violate any of the constraints in $\mathfrak{C}$.

\subsubsection*{Local optimization worker (\texttt{local\_opt\_worker})}	\label{sec:LocalOptimizationWorker}

The local optimization worker receives a candidate solution, $X$, which satisfies both constraints (\ref{eq:JobAdmissibility})--(\ref{eq:RestWidthAdmissibility}), i.e., job-admissibility and rest-width constraint. The objective is to perform local \emph{operations} in the candidate solution $X$ under the \texttt{better\_reply} mode, so that the resulting assignment reduces the cost (\ref{eq:ObjectiveFunction}), while maintaining admissibility. 

The operation of the local-optimization worker is governed by $N_{\rm loc}\in\mathbb{N}$ which is the total number of operations executed per run of this worker. A description of this worker is provided in Table~\ref{Tb:LocalOptWorker}. Note that the number of steps executed within this worker is fixed and independent on whether a cost reduction is found or not.

\begin{table}[th!]
\boxed{\footnotesize{
\begin{minipage}{0.98\textwidth}
\begin{tabular}{ll}
\hline
\textbf{function} \texttt{$X$ = local\_opti\_worker($X$,$\mathfrak{C}$)} & \\\hline
%\texttt{\BLUE{RAND SET} t from [0,1]} & \\
%\texttt{\BLUE{IF} ( t > $\beta$ ) } & \\
\texttt{\BLUE{REPEAT} (for $N_{\rm loc}$ times) } & \\
\quad \texttt{\BLUE{RAND SET} \RED{oper} from \{1,2,...,|$\mathcal{O}$|\}} & \\
\quad \texttt{$X=\mathcal{O}$[\RED{oper}].better\_reply($X$,$\mathfrak{C}$)} & \\
\texttt{\BLUE{END REPEAT}} & \\
%\texttt{\BLUE{ELSE} } & \\
%\quad \texttt{\BLUE{REPEAT} ($N_{\rm rnd}$ times) } & \\
%\quad \quad \texttt{\BLUE{RAND SET} oper from \{1,2,...,|$\mathcal{O}$|\}} & \\
%\quad \quad \texttt{$X'$ = operations[oper].random\_search($X$);} & \\
%%\quad \quad \texttt{\BLUE{IF} ($X'$ admissible and $g(X')$ $\leq$ $g(X)$) } & \\
%%\quad \quad \quad \texttt{replace $X$ with $X'$} & \\
%%\quad \quad \texttt{\BLUE{END IF}} & \\
%\quad \texttt{\BLUE{END REPEAT}} & \\
%\texttt{\BLUE{END IF}} & \\
\end{tabular}
\end{minipage}
}
}
\caption{\texttt{local\_opti\_worker}}
\label{Tb:LocalOptWorker}
\end{table}

\subsubsection*{Perturbation worker (\texttt{perturb\_worker})}	\label{sec:PerturbationWorker}

With probability $\lambda>0$, a number of operations under the \texttt{random\_reply} mode are executed while maintaining admissibility and without necessarily reducing the cost (\ref{eq:ObjectiveFunction}). A fixed number of $N_{\rm per}$ operations is executed per each run of this worker. The goal of such operations is to escape from local optima by temporarily accepting worse performances. A description of the architecture of the perturbation worker is provided in Table~\ref{Tb:PerturbWorker}.

\begin{table}[tbh!]
\boxed{\footnotesize{
\begin{minipage}{0.98\textwidth}
\begin{tabular}{ll}
\hline
\textbf{function} \texttt{$X$ = perturb\_worker($X$,$\mathfrak{C}$)} & \\\hline
\texttt{\BLUE{RAND SET} t in [0,1]} & \\
\texttt{\BLUE{IF} ( t < $\lambda$ ) } & \\
\quad \texttt{\BLUE{REPEAT} (for $N_{\rm per}$ times) } & \\
\quad \quad \texttt{\BLUE{RAND SET} \RED{oper} from \{0,1,...,$\magn{\mathcal{O}}$\}} & \\
\quad \quad \texttt{$X=\mathcal{O}$[\RED{oper}].random\_reply($X$,\{$\mathfrak{C}_{\rm job}$,$\mathfrak{C}_{\rm rw}$\})} & \\
\quad \texttt{\BLUE{END REPEAT}} & \\
\texttt{\BLUE{END IF}} & \\
\end{tabular}
\end{minipage}
}
}
\caption{\texttt{perturb\_worker}}
\label{Tb:PerturbWorker}
\end{table}

\subsection{Filtering (\texttt{filter})}	\label{sec:Filter}

The role of the \texttt{filter} is to assess the quality of the produced candidate solutions and to decide which of the candidate solutions need to be reprocessed and which should be reserved for later use. 

The operation of the \texttt{filter} is based upon the notions of \emph{reprocessing} and \emph{reservation}. In particular, the filter is responsible for updating two sets of candidate solutions, namely the \emph{main pool} $\mathfrak{P}$ and the \emph{reserve pool} $\mathfrak{R}$.
The pool, $\mathfrak{P}$, is the set of candidate solutions that are currently getting processed (i.e., they go through the optimization steps in \texttt{optimize}). The reserve pool, $\mathfrak{R}$, maintains candidate solutions from earlier stages of the optimization of \emph{high potential} that may replace candidate solutions within the main pool $\mathfrak{P}$ upon request.  

More specifically, the main steps of the \texttt{filter} are shown in Figure~\ref{fig:architecture}. In words, for each candidate solution $X_i$ of the main pool $\mathfrak{P}$, we first assess its \emph{good standing} and decide on whether it should further be processed based on prior performances summarized in memory $\mathfrak{M}(X_i)$. In particular, if $X_i$ is \emph{not} of a good standing, then it is removed from the main pool (i.e., \texttt{$\mathfrak{P}$.remove($X_i$)}) and replaced by another candidate solution from the reserve pool (i.e., \texttt{$X_r=\mathfrak{R}$.get()} and \texttt{$\mathfrak{P}$.add($X_r$)}). If, instead, $X_i$ is of a good standing, then it remains part of the main pool $\mathfrak{P}$ and continues on the assessment of its \emph{high potential}, based again on prior performances summarized in memory $\mathfrak{M}(X_i)$. If $X_i$ is identified as having high potential, then it is saved into the reserve pool $\mathfrak{R}$ (i.e., \texttt{$\mathfrak{R}$.add($X_i$)}). When all candidate solutions have been filtered (i.e., assessed with respect to their good standing and high potential), then the main pool $\mathfrak{P}$ exits the \texttt{filter}.

The reserve pool $\mathfrak{R}$ is initially empty and it is gradually populated by solutions of high-potential. However, we assume that $\mathfrak{R}$ has a bounded maximum size $\magn{\mathfrak{R}}_{\max}$, which means that when its capacity is reached, the newly entered solution will replace the oldest one in the pool. The reason for selecting a bounded maximum size is to allow for dynamic restarts within a bounded fitness-distance from the current best. In many practical scenarios, performing dynamic restarts from very early stages may delay significantly the optimization speed. However, by properly selecting the maximum size of the reserve pool, we may find an appropriate balance between earlier dynamic restarts and optimization speed. 

It is also important to note that the maximum size of the main pool $\mathfrak{P}$ is also bounded by $\magn{\mathfrak{P}}_{\max}$ (as discussed in Section~\ref{sec:Initialization}). When a reserved solution needs to replace another one from the main pool, the oldest one in $\mathfrak{R}$ is always picked. Also, when a reserved solution is requested to replace $X\in\mathfrak{P}$ but $\mathfrak{R}$ is currently empty, then a replacement is not possible and the size of the main pool is going to reduce by one. When the main pool $\mathfrak{P}$ becomes empty, then the optimization should terminate.

%\begin{table} [ht!]
%\boxed{\footnotesize{
%\begin{minipage}{0.98\textwidth}
%\begin{tabular}{ll}
%\hline
%\textbf{function} \texttt{\BLUE{BOOL} filter($X$)} & \\\hline
%\texttt{//assess good\_standing} & \\
%%\texttt{\BLUE{SET} gs = good\_standing($X$,$\mathfrak{M}(X)$,$\hat{X}$);} & \mbox{(assess good-standing)} \\
%\texttt{\BLUE{IF} (cont\_processing(gs,$X$,$\mathfrak{M}(X)$) == \BLUE{TRUE})} & \mbox{(if continue processing)} \\
%\quad \texttt{$\mathfrak{P}$.add($X$);} & \mbox{(add to the pool)} \\
%\texttt{\BLUE{ELSE} } & \\
%\quad \texttt{$X_r$ = $\mathfrak{R}$.get();} & \mbox{(get reserved solution)} \\
%\quad \texttt{$\mathfrak{P}$.add($X_r$);} & \mbox{(add to the pool)} \\
%\texttt{\BLUE{END IF}} & \\
%\hline
%\texttt{//assess high\_potential, add\_to\_reserves} & \\
%\texttt{\BLUE{SET} hp = high\_potential($X$,$\mathfrak{M}(X)$,$\hat{X}$);} & \mbox{(assess high-potential)} \\
%\texttt{\BLUE{IF} (add\_to\_reserves(hp,$X$,$\mathfrak{M}(X)$) == \BLUE{TRUE})} & \\
%\quad \texttt{$\mathfrak{R}$.add($X$);} & \mbox{(add to the reserves)} \\
%\texttt{\BLUE{END IF}} & \\
%\hline
%\texttt{//check terminate criterion} & \\
%\texttt{\textcolor{blue}{if} (terminate($\mathfrak{P}$,timer.current())} & \\
%\quad \texttt{\textcolor{blue}{RETURN FALSE};} & \mbox{(terminate processing)} \\
%\texttt{\BLUE{ELSE}} & \\
%\quad \texttt{\BLUE{RETURN TRUE};} & \mbox{(continue processing)} \\
%\texttt{\BLUE{END IF}} & \\
%\end{tabular}
%\end{minipage}
%}
%}
%\caption{\texttt{filter()}}
%\label{Tb:Filter}
%\end{table}

\subsubsection*{Good standing (\texttt{good\_standing})}		\label{sec:GoodStanding}

The function \texttt{good\_standing} assesses the potential of a candidate solution $X$ to continue providing reductions in the cost value. It depends on the current version of the candidate solution $X$, its prior memory $\mathfrak{M}(X)$, and the currently best candidate solution $\hat{X}$. Informally, the role of the good-standing assessment is to capture whether a candidate solution $X$ is currently able to follow the path of the currently best candidate $\hat{X}$, and this is achieved by evaluating a) the fitness-distance of $X$ from the current best $\hat{X}$, b) the steps elapsed since the last time $X$ provided an improvement to $\hat{X}$, and c) the gradient of the cost reduction during the most recent processing steps of $X$. In fact, we would like that the candidate solution $X$ is sufficiently close to the current best, and its cost gradient is sufficiently large.

To accomplish this assessment, we first introduce the following parameters: 
\begin{itemize}
\item $N_{\rm gs}^*$ denotes the number of processing steps over which the good standing of a candidate solution is evaluated.
\item $D(X,\hat{X})$ denotes the normalized fitness-distance between $X$ and the current best $\hat{X}$, defined as follows:
$$D(X,\hat{X}) \df \frac{g(X)-g(\hat{X})}{g(\hat{X})}.$$
\item $D_{\rm gs}^*$ denotes the maximum allowed normalized fitness distance from the current best for which the good standing is maintained.
\item $G(X,\mathfrak{M}(X),N_{\rm gs}^*)$ denotes the gradient of cost reduction observed in the candidate solution $X$, defined as follows:
$$G(X,\mathfrak{M}(X),N_{\rm gs}^*) \df \frac{g(X[N_{\rm ps}-N_{\rm gs}^*]) - g(X[N_{\rm ps}])}{N_{\rm gs}^* \cdot g(X[N_{\rm ps}-N_{\rm gs}^*])} \geq 0,$$
for $N_{\rm ps} > N_{\rm gs}^*$, where $N_{\rm gs}^*$ denotes the number of recent processing steps over which we estimate the gradient change of the cost function. % Recall also that $N_{\rm ps}=N_{\rm ps}(X)$ corresponds to the total number of processing steps of $X$.
% \item $N_{\rm tps}^{0}$ is the total number of processing steps after which we check whether a candidate solution $X$ is rather far from the current best $\hat{X}$.
\item $G_{\rm gs}^*$ denotes a threshold based on which the gradient $G$ of cost reduction is being evaluated.
\item $N_{\rm rb}(X)$ denotes the number of processing steps since the candidate solution path leading to $X$ has provided a \emph{recent best} (i.e., a reduction to the cost function).
% \item $N_{\rm rb}^*$ denotes the number of prior processing steps that constitutes a \emph{recent} processing history.
\end{itemize}

In particular, the exact steps for assessing the good standing of $X$ are provided in Table~\ref{Tb:GoodStanding}.
\begin{table} [ht!]
\boxed{\footnotesize{
\begin{minipage}{0.98\textwidth}
\begin{tabular}{ll}
\hline
\textbf{function} \texttt{\BLUE{BOOL} good\_standing($X$,$\mathfrak{M}(X)$,$\hat{X}$)} & \\\hline
\texttt{\BLUE{IF} ( $N_{\rm ps}(X)$ > $N_{\rm gs}^*$) } & \mbox{(if enough processing steps)} \\
\quad \texttt{\BLUE{IF} ($D(X,\hat{X})$ < $D_{\rm gs}^*$ {and} $N_{\rm rb}(X)$ < $N_{\rm gs}^*$) } & \mbox{(if $X$ is ``close'' to current best and } \\
& \mbox{recently provided an improvement)} \\
\quad \quad \texttt{\BLUE{IF} ($G(X,\mathfrak{M}(X))$ > $G_{\rm gs}^*$) } & \mbox{(if progress rate is large)}\\
\quad \quad \quad \texttt{\BLUE{RETURN TRUE}} & \\
\quad \quad \texttt{\BLUE{ELSE}} & \mbox{(if progress rate is small)} \\
\quad \quad \quad \texttt{\BLUE{RETURN FALSE}} & \\
\quad \quad \texttt{\BLUE{END IF}} & \\
\quad \texttt{\BLUE{ELSE}} & \mbox{(if $X$ is ``far'' from current best or} \\
& \mbox{recently provided no improvement)} \\
\quad \quad \texttt{\BLUE{RETURN FALSE}} & \\
\quad \texttt{\BLUE{END IF}} & \\
\texttt{\BLUE{ELSE}} & \mbox{(if not enough processing steps yet)} \\
\quad \texttt{\BLUE{RETURN TRUE}} & \mbox{(good standing at early stages)} \\
\texttt{\BLUE{END IF}} & \\
\end{tabular}
\end{minipage}
}
}
\caption{\texttt{good\_standing}}
\label{Tb:GoodStanding}
\end{table}
First, we check whether enough processing steps have elapsed for the good standing criterion to be evaluated (i.e., $N_{\rm ps}(X) \geq N_{\rm gs}^*$). This initial check is required due to the absence of any prior knowledge regarding the potential of a candidate solution $X$. Then, we evaluate a) the normalized fitness-distance of $X$ from the current best $\hat{X}$, $D(X,\hat{X})$, and b) the processing steps elapsed since the last improvement offered by $X$, $N_{\rm rb}(X)$. Both have to be small enough to maintain a good standing. Lastly, we check whether $X$ exhibits a sufficiently large progress rate. If its progress rate is larger than $G_{\rm gs}^*$, then $X$ will retain its good standing. In other words, for maintaining a good standing, it is not sufficient that $X$ is close enough (with respect to fitness) to the current best, rather it should also provide a progress rate that is promising for improving the current best.

The parameters $D_{\rm gs}^*$, $G^*$, and $N_{\rm gs}^*$ need to be determined by the user.

\subsubsection*{High potential (\texttt{high\_potential})}	\label{sec:HighPotential}

The notion of \emph{high potential} of a candidate solution $X$ captures its ability to provide significant reductions to the objective function (\ref{eq:ObjectiveFunction}). For example, an indicative criterion of a high potential is an extremely high progress rate $G(X,\mathfrak{M}(X))$. We wish to store candidate solutions exhibiting high potential at different stages of their processing paths, so that we are able to \emph{restart} the processing from these stages. This form of restarts may serve as a mechanism for escaping from local optima, while at the same time it may increase the probability of converging to the optimal solution.

To assess the high potential of a candidate solution, we first introduce the following notation:
\begin{itemize}
\item $N_{\rm hp}^*$ denotes the number of processing steps over which the high potential of a candidate solution is evaluated.
\item $D_{\rm hp}^*$ denotes the maximum allowed normalized fitness-distance from the current best for which the high potential property can be assessed.
%\item $H(X,\mathfrak{M}(X))$ denotes the gradient of cost reduction observed during the last $N_{\rm hp}^*$ number of processing steps of $X$, defined as follows:
%\begin{equation*}
%G(X,\mathfrak{M}(X),N_{\rm hp}^*) \df \frac{g(X[N_{\rm ps} - N_{\rm hp}^*]) - g(X[N_{\rm ps}])}{N_{\rm hp}^*\cdot g(X[N_{\rm ps}-N_{\rm hp}^*])}.
%\end{equation*}
%where recall that $N_{\rm ps}$ is the number of processing steps of the candidate solution $X$;
\item $G_{\rm hp}^*$ denotes a threshold based on which the gradient of cost reduction $G$ is being evaluated.
\item $N_{\rm lr}(X)$ denotes the elapsed processing steps since the candidate solution $X$ was last reserved into $\mathfrak{R}$.
% \item $N_{\rm lr}^*$ denotes a threshold of elapsed steps since a candidate solution was last reserved below which no reservation is allowed.
\end{itemize}

The exact steps for assessing the high potential of a candidate solution $X$ are provided in Table~\ref{Tb:HighPotential}. First we check whether enough processing steps have elapsed for the high-potential criterion to be evaluated. Then, we evaluate a) the normalized fitness-distance of $X$ from the current best $\hat{X}$, $D(X,\hat{X})$, and b) the processing steps elapsed since the last reservation offered by the processing path of $X$. We would like $X$ to be close enough to the current best, however we would also like $X$ not to have provided recently any reservations (thus not creating subsequent reservations from the same processing paths). Lastly, we check whether $X$ exhibits a sufficiently large progress rate, i.e., larger than $G_{\rm hp}^*$. 
\begin{table} [ht!]
\boxed{\footnotesize{
%\begin{minipage}{0.98\textwidth}
%\begin{tabular}{ll}
%\texttt{\textcolor{blue}{if} ($N_{\rm ps}(X)$>$0$ \&\& $N_{\rm tps}$ > $N_{\rm tps}^0$) \{ } & \mbox{(check processing steps)} \\
%\quad \texttt{\textcolor{blue}{if} ($D(X,\hat{X})$ > $D_{\rm hp}^*$) } & \mbox{(check distance from current best)} \\
%\quad \quad \texttt{\textcolor{blue}{return false}; } & \\
%\texttt{\}} & \\
%\texttt{\textcolor{blue}{else} } & \\
%\quad \texttt{\textcolor{blue}{return true}; } & \\
%\texttt{\textcolor{blue}{if} ($N_{\rm ps}(X)$ < $N_{\rm hp}^*$) } & \mbox{(check if history is sufficiently large)} \\
%\quad \texttt{\textcolor{blue}{return false}; } & \\
%\texttt{\textcolor{blue}{else} \{ } & \\
%\quad \quad \texttt{\textcolor{blue}{if} ($H(X,\mathfrak{M}(X))$ > $H^*$)} & \\
%\quad \quad \quad \texttt{\textcolor{blue}{return true};} & \\
%\quad \quad \texttt{\textcolor{blue}{else}} & \\
%\quad \quad \quad \texttt{\textcolor{blue}{return false};} & \\
%\texttt{\}} & \\
%\end{tabular}
%\end{minipage}
\begin{minipage}{0.98\textwidth}
\begin{tabular}{ll}
\hline
\textbf{function} \texttt{\BLUE{BOOL} high\_potential($X$,$\mathfrak{M}(X)$,$\hat{X}$)} & \\\hline
\texttt{\BLUE{IF} ( $N_{\rm ps}(X)$ > $N_{\rm hp}^*$ ) } & \mbox{(if enough processing steps)} \\
\quad \texttt{\BLUE{IF} ($D(X,\hat{X})$ < $D_{\rm hp}^*$ and $N_{\rm lr}(X)$ > $N_{\rm hp}^*$) } & \mbox{(if $X$ is ``close'' to current best and } \\
& \mbox{has not recently been reserved)} \\
\quad \quad \texttt{\BLUE{IF} ($G(X,\mathfrak{M}(X))$ > $G_{\rm hp}^*$) } & \mbox{(if progress rate is large)}\\
\quad \quad \quad \texttt{\BLUE{RETURN TRUE}} & \\
\quad \quad \texttt{\BLUE{ELSE}} & \mbox{(if progress rate is small)} \\
\quad \quad \quad \texttt{\BLUE{RETURN FALSE}} & \\
\quad \quad \texttt{\BLUE{END IF}} & \\
\quad \texttt{\BLUE{ELSE}} & \mbox{(if $X$ is ``far'' from current best or} \\
& \mbox{it has recently been reserved)} \\
\quad \quad \texttt{\BLUE{RETURN FALSE}} & \\
\quad \texttt{\BLUE{END IF}} & \\
\texttt{\BLUE{ELSE}} & \mbox{(if not enough processing steps)} \\
\quad \texttt{\BLUE{RETURN TRUE}} & \mbox{(assume high potential at early stages)} \\
\texttt{\BLUE{END IF}} & \\
\end{tabular}
\end{minipage}
}
}
\caption{\texttt{high\_potential}}
\label{Tb:HighPotential}
\end{table}

% It also worth noting that due to the absence of any a-priori knowledge regarding the quality of the candidate solutions, at the early optimization stages we assume a high potential for all the available candidate solutions in $\mathfrak{P}$.

We should expect that $D_{\rm gs}^*>D_{\rm hp}^*$ and that $G_{\rm gs}^*<G_{\rm hp}^*$, since \emph{high-potential should also imply good-standing but not the other way around}.

\subsection{Selection (\texttt{select})}		\label{sec:Selection}

The last part of the optimization algorithm is the selection phase (\texttt{select}). In this phase, the main pool $\mathfrak{P}$ of candidate solutions has already been filtered, and the question is whether we should terminate processing and select the current best estimate $\hat{X}$ or continue processing $\mathfrak{P}$. The specifics of this phase are shown in Figure~\ref{fig:architecture}. The terminate decision is taken upon the current size of the main pool $\mathfrak{P}$ as well as the maximum allowed processing time $t_{\max}$ imposed by the user. In particular, when the size of the main pool is zero, or when the elapsed processing time has exceeded $t_{\max}$, then the optimization is terminated and the current best estimate $\hat{X}$ is provided as an output.

%%%%%%%%%%%%%%%%%%%%%%%%%%%%%%%%%%%%%%%%%%%%%%%%%%%%%%%%%%%%%%%%%%%%%%%%%%%%%%%%%%%%%%%%%%%%%%
\section{Operations}	\label{sec:Operations}

As we mentioned in the description of the optimization phase (\texttt{optimize}), the means by which local modifications of candidate solutions are created/searched is through a set of available operations $\mathcal{O}$. In this section, we would like to provide a description of the operations considered, as well as a description of their alternative modes (as initially introduced in Section~\ref{sec:Optimize}), namely \texttt{better\_reply}, \texttt{constr\_reply} and \texttt{random\_reply}. 

The operations implemented are the following:
\begin{itemize}
\item \texttt{MoveItem},
%\item \texttt{SwitchObject},
\item \texttt{SwapItems},
\item \texttt{SplitItem},
\item \texttt{RemoveObject},
\item \texttt{ReverseRemoveObject},
\item \texttt{RemoveItem},
\end{itemize}

% In the following section we provide a detailed description of the \texttt{better\_reply}() assigned to each one of the above operations. 

%In the following section we provide a detailed description of the two main modes (\texttt{random\_init()} and \texttt{better\_reply}()) assigned to each one of the above operations. 
%
%The description of the operations will be based on the following notation:
%\begin{itemize}
%\item $N_{\rm rt}$ denotes the number of random trials exercised during the random initialization of an operation;
%\end{itemize}

\subsection{\texttt{MoveItem} operation}	\label{sec:MoveItem}

\begin{figure}[h!]
\centering
\includegraphics[scale=1]{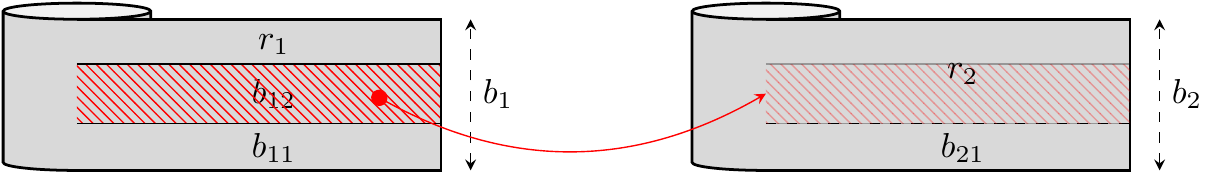}
\caption{\texttt{MoveItem} operation.}
\label{fig:move_operation}
\end{figure}

The move operation moves a single item from one object (or roll) to another object (see Figure~\ref{fig:move_operation}). The details of the \texttt{better\_reply} mode of this operation are described in Table~\ref{Tb:MoveItemBetterReply}.
\begin{table} [ht!]
\boxed{\small{
\begin{minipage}{0.98\textwidth}
\begin{tabular}{ll}
\textbf{function} \texttt{$X=$ MoveItem.better\_reply($X$,$\mathfrak{C}$,$g()$)} & \\\hline
\texttt{\BLUE{SET} used\_objects = get\_used\_objects($X$);} & \\
\texttt{\BLUE{REPEAT} (for $N_{\rm br\_trials}$ times) } & \\
\quad \texttt{\BLUE{RAND SET} \VIOLET{object1} from used\_objects} & \\
% \quad \texttt{\BLUE{SET} items1 = all items of object1} & \\
\quad \texttt{\BLUE{FOR} (each \VIOLET{item} of \VIOLET{object1}) } & \\
%\quad \quad \texttt{\BLUE{RAND SET} \VIOLET{item} from items1;} &\\
\quad \quad \texttt{\BLUE{REPEAT} (for $N_{\rm br\_trials}$ times) } & \\
\quad \quad \quad \texttt{\BLUE{RAND SET} \VIOLET{object2} from all objects but \VIOLET{object1}} & \\
\quad \quad \quad \texttt{$X'$ = $X$:move \VIOLET{item} from \VIOLET{object1} to \VIOLET{object2}} & \\
\end{tabular}
\fbox{\colorbox{lightgray}{
\begin{minipage}[fill=gray]{0.96\textwidth}
\begin{tabular}{ll}
\quad \quad \texttt{\textcolor{blue}{IF} ($X'$ is $\mathfrak{C}$-admissible and $g(X')$ < $g(X)$) } & \rdelim\}{3}{20pt}[\textit{\scriptsize better reply}] \\
\quad \quad \quad \texttt{\BLUE{SET} $X$ = $X'$ and \BLUE{RETURN} $X$} & \\
\quad \quad \texttt{\BLUE{END IF}} & \\
\end{tabular}
\end{minipage}}}
\begin{tabular}{ll}
\quad \quad \texttt{\BLUE{END FOR}} & \\
\quad \texttt{\BLUE{END FOR}} & \\
\texttt{\BLUE{END REPEAT}} & \\
\texttt{\BLUE{RETURN} $X$} & \\
\end{tabular}
\end{minipage}
}
}
\caption{\texttt{MoveItem.better\_reply}}
\label{Tb:MoveItemBetterReply}
\end{table}
Note that this mode implements a form of nested search over the source object (\texttt{object1}), an item from this object (\texttt{item}) and the destination object (\texttt{object2}). Then, it operates a move of the \texttt{item} from \texttt{object1} to \texttt{object2}. The \texttt{object1} should be selected from the set of objects that are currently used (\texttt{used\_objects}), i.e., from the set of objects with at least one item assigned to them. Given that the number of objects might be large, we can bound the size of this nested search via the parameter $N_{\rm br\_trials}$. The selection of this parameter depends on the number of available objects $\mathcal{J}$, and the desired time of optimization $t_{\max}$ determined by the user.

Similar in architecture is the \texttt{constr\_reply} mode of this operation, however the objective is to simply generate admissible assignments for those objects that do not satisfy the constraint. The details are shown in Table~\ref{Tb:MoveItemConstrReply}.
\begin{table} [ht!]
\boxed{\small{
\begin{minipage}{0.98\textwidth}
\begin{tabular}{ll}
\textbf{function} \texttt{$X=$ MoveItem.constr\_reply($X$,\{$\mathcal{C}_{\rm job},\mathcal{C}_{\rm rw}$\},$\mathcal{B}$)} & \\\hline
\texttt{\BLUE{REPEAT} ($N_{\rm con\_trials}$ times) } & \\
\quad \texttt{\BLUE{RAND SET} \VIOLET{bad\_object} from $\mathcal{B}$} & \\
% \quad \texttt{\BLUE{SET} items1 = all items of object1} & \\
\quad \texttt{\BLUE{FOR} (each \VIOLET{item} of \VIOLET{bad\_object}) } & \\
%\quad \quad \texttt{\BLUE{RAND SET} \VIOLET{item} from items1;} &\\
\quad \quad \texttt{\BLUE{REPEAT} ($N_{\rm con\_trials}$ times) } & \\
\quad \quad \quad \texttt{\BLUE{RAND SET} \VIOLET{object2} from all objects but \VIOLET{bad\_object}} & \\
\quad \quad \quad \texttt{$X'$ = $X$:move \VIOLET{item} from \VIOLET{bad\_object} to \VIOLET{object2}} & \\
\end{tabular}
\fbox{\colorbox{lightgray}{
\begin{minipage}[fill=gray]{0.96\textwidth}
\begin{tabular}{ll}
\quad \quad \texttt{\BLUE{IF} (\VIOLET{item} is $\mathcal{C}_{\rm job}$-admissible and \VIOLET{bad\_object} is $\mathcal{C}_{\rm rw}$-admissible) } & \\
%\rdelim\}{3}{20pt}[\textit{\scriptsize $\mathfrak{C}$-admissibility}] \\
\quad \quad \quad \texttt{\BLUE{SET} $X$ = $X'$ and \BLUE{RETURN} $X$} & \\
\quad \quad \texttt{\BLUE{END IF}} & \\
\end{tabular}
\end{minipage}}}
\begin{tabular}{ll}
\quad \quad \texttt{\BLUE{END REPEAT}} & \\
\quad \texttt{\BLUE{END FOR}} & \\
\texttt{\BLUE{END REPEAT}} & \\
\texttt{\BLUE{RETURN} $X$} & \\
\end{tabular}
\end{minipage}
}
}
\caption{\texttt{MoveItem.constr\_reply}}
\label{Tb:MoveItemConstrReply}
\end{table}
In particular, for $N_{\rm con\_trials}$ times, we go through the set of ``bad'' objects that do not satisfy the constraint. For each one of the items of this object, we go through a number of destination objects (different than the ``bad'' object), and we check whether admissibility of the modified assignment has been resolved with respect to the ``bad'' object (without violating the job-admissibility requirements for the moved item). Similarly to the \texttt{better\_reply} mode, this mode is controlled by appropriately selecting the parameter $N_{\rm con\_trials}$.

The \texttt{random\_reply} mode of this operation is described in Table~\ref{Tb:MoveItemRandomInit}. 
\begin{table} [ht!]
\boxed{\small{
\begin{minipage}{0.98\textwidth}
\begin{tabular}{ll}
\textbf{function} \texttt{$X=$ MoveItem.random\_init($X$,$\mathfrak{C}$)} & \\\hline
\texttt{\BLUE{SET} \VIOLET{used\_objects} = get\_used\_objects($X$);} & \\
\texttt{\BLUE{REPEAT} (for $N_{\rm rand\_trials}$ times) } & \\
\quad \texttt{\BLUE{RAND SET} \VIOLET{object1} from \VIOLET{used\_objects}} & \\
\quad \texttt{\BLUE{RAND SET} \VIOLET{object2} from \VIOLET{used\_objects}} & \\
\quad \texttt{\BLUE{IF} (\VIOLET{object1} is not equal to \VIOLET{object2}) } & \\
% \quad \quad \texttt{\BLUE{SET} \VIOLET{items1} = all items of \VIOLET{object1}} & \\
\quad \quad \texttt{\BLUE{RAND SET} \VIOLET{item} from the items of \VIOLET{object1}} & \\
\quad \quad \texttt{$X'$ = $X$:move \VIOLET{item} from \VIOLET{object1} to \VIOLET{object2}} & \\
\end{tabular}\\
\fbox{\colorbox{lightgray}{
\begin{minipage}[fill=gray]{0.96\textwidth}
\begin{tabular}{ll}
\quad \quad \texttt{\textcolor{blue}{IF} ($X'$ is $\mathfrak{C}$-admissible) } & \rdelim\}{3}{20pt}[\textit{\scriptsize $\mathfrak{C}$-admissibility}] \\
\quad \quad \quad \texttt{\BLUE{SET} $X$ = $X'$ and \BLUE{RETURN} $X$} & \\
\quad \quad \texttt{\BLUE{END IF}} & \\
\end{tabular}
\end{minipage}}}
\begin{tabular}{ll}
\quad \texttt{\BLUE{END IF}} & \\
\texttt{\BLUE{END REPEAT}} & \\
\texttt{\BLUE{RETURN} $X$} & \\
%\texttt{\}} & \\
\end{tabular}
\end{minipage}
}
}
\caption{\texttt{MoveItem.random\_init}}
\label{Tb:MoveItemRandomInit}
\end{table}
It simply relies on a random selection of two (distinct) objects, one of which should have a non-zero number of items assigned to it. Then, an item is randomly picked from one of the objects and is assigned to the second object. The objective is to generate a new assignment $X'$ that maintains $\mathfrak{C}$-admissibility. The effect of this mode into the cost is not considered. 

Finally, note that there is not a unique way to initialize the objects/items involved in these modes, and this is also true for all the operations that follow. For example, in \texttt{better\_reply} mode, we have selected a nested type of search to initialize the objects/items involved, while in \texttt{random\_reply} mode, we implement a completely random initialization. This is not restrictive. In practice, a nested type of initialization of the parameters provided a faster discovery of better replies, however formulating the most efficient architecture for these modes requires a separate investigation.

\subsection{\texttt{SwapItems} operation}	\label{sec:SwapItems}

\begin{figure}[h!]
\centering
\includegraphics[scale=1]{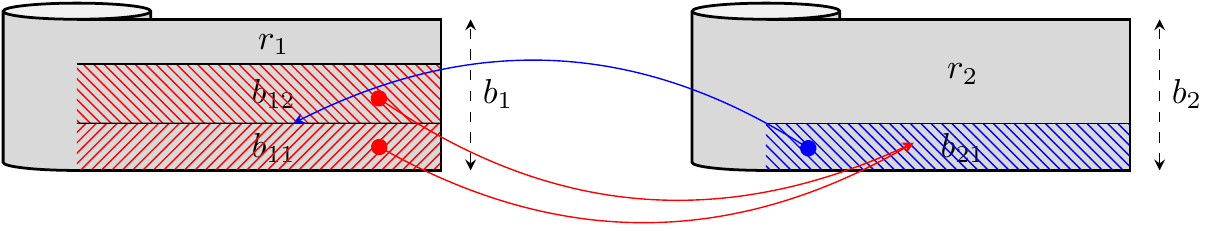}
\caption{\texttt{SwapItems} operation.}
\label{fig:SwapItems_operation}
\end{figure}

The \texttt{SwapItems} operation swaps one or more items from one object (or roll) with one or more items from another object (see Figure~\ref{fig:SwapItems_operation}). %The difference with the \emph{switch object} operation is the fact that the swap operation requires the existence of items in both objects. Furthermore, the \texttt{SwapItems} operation may involve more than two objects. 

In particular, the \texttt{SwapItems} operation is based upon the selection of two objects (among all objects with non-zero items) and the selection of a combination of items from these objects. Then, the selected items from both objects are swapped with each other. Figure~\ref{fig:SwapItems_operation} provides an example, where two items from object $1$ are swapped with one item from object $2$. % The number of objects and the number of items involved defines the \textit{type} of the swap operation. The types considered depend on the physics of the problem itself. For example, when the maximum number of items that may fit into an object is $3$, it is natural that only swap operations that involve combinations of up to 3 items are considered. 

%\begin{table} [ht!]
%\boxed{\footnotesize{
%\begin{minipage}{0.98\textwidth}
%\begin{tabular}{ll}
%\texttt{used\_objects = get\_used\_objects($X$);} & \\
%\texttt{\textcolor{blue}{for} (\textcolor{blue}{int} i=0; i<$N_{\rm rand\_trials}$; i++) \{ } & \\
%\quad \texttt{\textcolor{blue}{int} object1 = rand(used\_objects);} & \\
%\quad \texttt{\textcolor{blue}{int} object2 = rand(used\_objects);} & \\
%\quad \texttt{\textcolor{blue}{if} (object1 != object2) \{ } & \\
%\quad \quad \texttt{\textcolor{green}{//Computing combinations of items of object 1} } & \mbox{} \\
%\quad \quad \texttt{items1 = get\_items($X$,object1);} & \mbox{(items from object 1)} \\
%\quad \quad \texttt{type1 = rand(\{$1$,$2$,...,$N$\});} & \mbox{} \\
%\quad \quad \texttt{combs1 = get\_combs(items1,1,min\{|items1|,type1\});} & \mbox{} \\
%\quad \quad \texttt{\textcolor{green}{//Computing combinations of items of object 2} } & \mbox{} \\
%\quad \quad \texttt{items2 = get\_items($X$,object2);} & \mbox{(items from object 2)} \\
%\quad \quad \texttt{type2 = rand(\{$1$,$2$,...,$N$\});} & \mbox{} \\
%\quad \quad \texttt{combs2 = get\_combs(items2,1,min\{|items2|,type2\});} & \mbox{} \\
%\quad \quad \texttt{\textcolor{green}{//Executing swap operation} } & \mbox{} \\
%\quad \quad \texttt{$X'$ = execute($X$, items1, items2, object1, object2);} & \mbox{(execute operation)} \\
%\quad \quad \texttt{\textcolor{blue}{break};} & \\
%%\quad \texttt{\}} & \\
%%\texttt{\}} & \\
%\end{tabular}
%\end{minipage}
%}
%}
%\caption{\texttt{SwapItems.random\_init($X$)}}
%\label{Tb:SwapItemsRandom}
%\end{table}

The \texttt{better\_reply} mode of this operation, for the case of 2 objects, is described in Table~\ref{Tb:MoveItemBetterReply}. 
\begin{table} [ht!]
\boxed{{\footnotesize
\begin{minipage}{0.98\textwidth}
\begin{tabular}{ll}
\hline
\textbf{function} \texttt{$X=$ SwapItems.better\_reply($X$,$\mathfrak{C}$,$g()$)} & \\\hline
\texttt{\BLUE{SET} \VIOLET{used\_objects} = objects with non-zero items in $X$} & \\
\texttt{\BLUE{REPEAT} (for $N_{\rm br\_trials}$ times) } \\
\quad \texttt{\BLUE{RAND SET} \VIOLET{object1} from \VIOLET{used\_objects}} &  \\
%\quad \texttt{\BLUE{SET} \VIOLET{items1} = all items of \VIOLET{object1}} & \mbox{} \\
%\quad \texttt{\BLUE{RAND SET} \VIOLET{type1} from the set \{1,2,...,|\VIOLET{items1}|\}} & \\
%\quad \texttt{\BLUE{SET} \VIOLET{combs1} = all possible combinations of distinct items in \VIOLET{object1} } &  \\
\quad \texttt{\BLUE{FOR} (each combination \VIOLET{comb1} of items in \VIOLET{object1})} & \\
\quad \quad \texttt{\BLUE{REPEAT} (for $N_{\rm br\_trials}$ times) } & \\
\quad \quad \quad \texttt{\BLUE{RAND SET} \VIOLET{object2} from \VIOLET{used\_objects} but \VIOLET{object1}} & \\
% \quad \quad \quad \texttt{\BLUE{SET} \VIOLET{items2} = all items of \VIOLET{object2}} & \\
% \quad \quad \quad \texttt{\BLUE{RAND SET} \VIOLET{type2} from the set \{1,2,...,|\VIOLET{items2}|\}} & \\
%\quad \quad \quad \texttt{\BLUE{SET} \VIOLET{combs2} = all possible combinations of distinct items in \VIOLET{object2}} & \\
\quad \quad \quad \texttt{\BLUE{FOR} (each combination \VIOLET{comb2} of items in \VIOLET{object2}) } & \\
\quad \quad \quad \quad \texttt{$X'$ = $X$: swap all items in \VIOLET{comb1} with all items in \VIOLET{comb2}} & \\
\end{tabular}
\fbox{\colorbox{lightgray}{\begin{minipage}[fill=gray]{0.96\textwidth}
\begin{tabular}{ll}
\quad \quad \quad \texttt{\textcolor{blue}{IF} ($X'$ is $\mathfrak{C}$-admissible and $g(X')$ < $g(X)$) } & \rdelim\}{3}{20pt}[\textit{\scriptsize better reply}] \\
\quad \quad \quad \quad \texttt{\BLUE{SET} $X$ = $X'$ and \BLUE{RETURN} $X$} & \\
\quad \quad \quad \texttt{\BLUE{END IF}} & \\
\end{tabular}
\end{minipage}}}
\begin{tabular}{ll}
\quad \quad \texttt{\BLUE{END REPEAT}} & \\
\quad \texttt{\BLUE{END FOR}} & \\
\texttt{\BLUE{END REPEAT}} & \\
\texttt{\BLUE{RETURN} $X$} & \\
\end{tabular}
\end{minipage}
}
}
\caption{\texttt{SwapItems.better\_reply}}
\label{Tb:SwapItemsBetterReply}
\end{table}
This mode incorporates the swap of a combination of items from object $1$ with another combination of items from object $2$ (different than object $1$). This mode can be augmented by alternative swap types, where more objects are involved. % One such alternative type may involve three objects (object $1$, object $2$ and object $3$), and two distinct combinations of items from object $1$ are swapped with a combination of items from object $2$ and object $3$, respectively.

The \texttt{constr\_reply} and \texttt{random\_reply} modes of the \texttt{SwapItems} operation are similar to the corresponding modes of the \texttt{MoveItem} operation, however following the structure of the swap operation. Accordingly, the \texttt{random\_reply} mode is based upon a random selection of objects, and a random selection of a combination of items from these objects.

%------------------------------------------------------
\subsection{\texttt{SplitItem} operation}	\label{sec:SplitItem}

\begin{figure}[h!]
\centering
\includegraphics[scale=1]{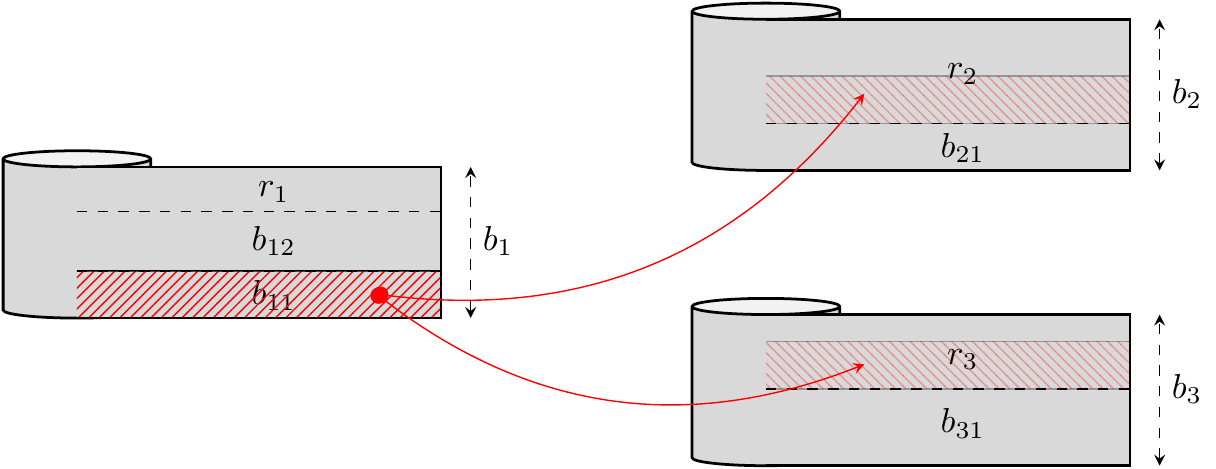}
\caption{\texttt{SplitItem} operation.}
\label{fig:SplitItem_operation}
\end{figure}

The \texttt{SplitItem} operation splits one item currently allocated in one object into two other distinct objects (see Figure~\ref{fig:SplitItem_operation}). %The \texttt{random\_reply()} mode of this operation simply relies on a random selection of both the involved objects and the involved item. In particular, an object is randomly picked from the set of used objects, and an item from that object is selected. Then, two different objects are randomly selected (used or unused), and  the selected item is assigned to both of these objects. % The formulation of the \texttt{random\_reply()} mode follows similar setup with the previous operations (see, e.g., Table~\ref{Tb:SwitchObjectRandomInit}).

%
%\begin{table} [ht!]
%\boxed{\footnotesize{
%\begin{minipage}{0.98\textwidth}
%\begin{tabular}{ll}
%\texttt{used\_objects = get\_used\_objects($X$);} & \\
%\texttt{\textcolor{blue}{for} (\textcolor{blue}{int} i=0; i<$N_{\rm rand\_trials}$; i++) \{ } & \\
%\quad \texttt{\textcolor{blue}{int} object1 = rand(used\_objects);} & \\
%\quad \texttt{\textcolor{blue}{int} object2 = rand(objects);} & \\
%\quad \texttt{\textcolor{blue}{int} object3 = rand(objects);} & \\
%\quad \texttt{\textcolor{blue}{if} ( object1 != object2 \&\& object1 != object3 ) \{ } & \\
%\quad \quad \texttt{\textcolor{green}{//Randomly selecting one item from object 1} } & \mbox{} \\
%\quad \quad \texttt{items1 = get\_items($X$,object1);} & \mbox{(items from object 1)} \\
%\quad \quad \texttt{item = rand(items1);} & \mbox{} \\
%\quad \quad \texttt{\textcolor{green}{//Executing split operation} } & \mbox{} \\
%\quad \quad \texttt{$X'$ = execute($X$, item, object1, object2, object3);} & \mbox{(execute operation)} \\
%\quad \quad \texttt{\textcolor{blue}{return true};} & \\
%%\quad \texttt{\}} & \\
%%\texttt{\}} & \\
%\end{tabular}
%\end{minipage}
%}
%}
%\caption{\texttt{SplitItem.random\_reply($X$)}}
%\label{Tb:SplitItemRandom}
%\end{table}

The \texttt{better\_reply} mode of this operation is described in Table~\ref{Tb:SplitItemBetterReply}. %
\begin{table} [ht!]
\boxed{{\footnotesize
\begin{minipage}{0.98\textwidth}
\begin{tabular}{ll}
\hline
\textbf{function} \texttt{$X=$ SplitItem.better\_reply($X$,$\mathfrak{C}$,$g()$)} & \\\hline
\texttt{\BLUE{SET} \VIOLET{used\_objects} = objects with non-zero items in $X$} & \\
\texttt{\BLUE{SET} \VIOLET{combs\_objects} = all combinations of 2 distinct objects out of $\mathcal{I}$ } & \\
\texttt{\BLUE{REPEAT} (for $N_{\rm br\_trials}$ times)} & \\
\quad \texttt{\BLUE{RAND SET} \VIOLET{object1} from \VIOLET{used\_objects}} &  \\
%\quad \texttt{\BLUE{SET} \VIOLET{items1} = all items of \VIOLET{object1}}&  \\
\quad \texttt{\BLUE{REPEAT} (for $N_{\rm br\_trials}$ times)} & \\
\quad \quad \texttt{\BLUE{RAND SET} a combination \{\VIOLET{object2},\VIOLET{object3}\} from \VIOLET{combs\_objects}} & \\
\quad \quad \texttt{\BLUE{IF} (\VIOLET{object1} does not coincide with either \VIOLET{object2} or \VIOLET{object3}) } & \\
\quad \quad \quad \texttt{\BLUE{FOR} (each item in \VIOLET{object1}) } & \\
\quad \quad \quad \quad \texttt{$X'$ = $X$: split item to \VIOLET{object2} and \VIOLET{object3}} & \\
\end{tabular}
\fbox{\colorbox{lightgray}{\begin{minipage}[fill=gray]{0.96\textwidth}
\begin{tabular}{ll}
\quad \quad \quad \texttt{\textcolor{blue}{IF} ($X'$ is $\mathfrak{C}$-admissible and $g(X')$ < $g(X)$) } & \rdelim\}{3}{20pt}[\textit{\scriptsize better reply}] \\
\quad \quad \quad \quad \texttt{\BLUE{SET} $X$ = $X'$ and \BLUE{RETURN} $X$} & \\
\quad \quad \quad \texttt{\BLUE{END IF}} & \\
\end{tabular}
\end{minipage}}}
\begin{tabular}{ll}
\quad \quad \quad \texttt{\BLUE{END FOR}} & \\
\quad \quad \texttt{\BLUE{END IF}} & \\
\quad \texttt{\BLUE{END REPEAT}} & \\
\texttt{\BLUE{END REPEAT}} & \\
\texttt{\BLUE{RETURN} $X$} & \\
\end{tabular}
\end{minipage}
}
}
\caption{\texttt{SplitItem.better\_reply}}
\label{Tb:SplitItemBetterReply}
\end{table}
This mode randomly selects an object (\texttt{object1}) from the set of \texttt{used\_objects} and a combination of 2 distinct objects (\texttt{object2} and \texttt{object3}) that are different from \texttt{object1}. Then it assigns a randomly selected item from \texttt{object1} to both \texttt{object2} and \texttt{object3}. This selection of objects and an item is repeated for a fixed number of times controlled by $N_{\rm br\_trials}$ or until a better reply is found. 

The \texttt{constr\_reply} and \texttt{random\_reply} modes of the \texttt{SplitItem} operation are similar to the random mode of the \texttt{MoveItem} operation, however following the structure of the \texttt{SplitItem} operation.

%------------------------------------------------------
\subsection{\texttt{RemoveObject} operation}	\label{sec:RemoveObject}

\begin{figure}[h!]
\centering
\includegraphics[scale=1]{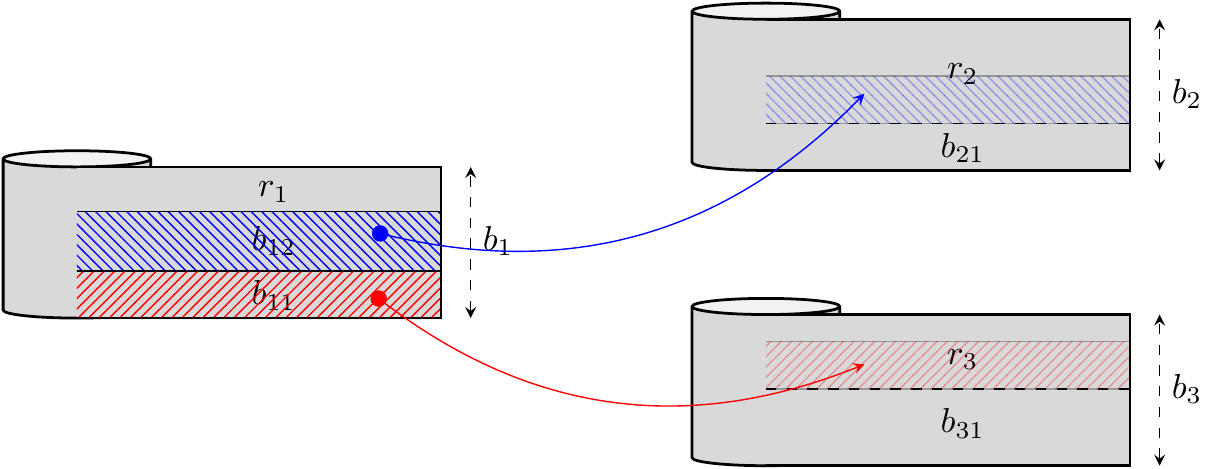}
\caption{\texttt{RemoveObject} operation.}
\label{fig:RemoveObject_operation}
\end{figure}

The \texttt{RemoveObject} operation involves an object with two or more items and moves a combination of its items to new distinct objects (which may or may not have items assigned to them), see Figure~\ref{fig:RemoveObject_operation}. In a way, the \texttt{RemoveObject} operation compliments the \texttt{MoveItem} operation, since it involves more than one item and more than one destination objects. 

% The \texttt{random\_reply()} of this operation simply relies on a random selection of \emph{source} and \emph{target} objects as well as the items to be moved. This mode follows similar structure with the corresponding modes of the previously described operations (see, e.g., Table~\ref{Tb:MoveItemRandomInit}).

The \texttt{better\_reply} mode of this operation is described in Table~\ref{Tb:RemoveObjectBetterReply}. 
\begin{table} [ht!]
\boxed{{\footnotesize
\begin{minipage}{0.98\textwidth}
\begin{tabular}{ll}
\hline
\textbf{function} \texttt{$X=$ RemoveObject.better\_reply($X$,$\mathfrak{C}$,$g()$)} & \\\hline
\texttt{\BLUE{SET} \VIOLET{used\_objects} = all objects with at least 2 items in $X$} & \\
\texttt{\BLUE{REPEAT} (for $N_{\rm br\_trials}$ times)} & \\
\quad \texttt{\BLUE{RAND SET} \VIOLET{object1} from \VIOLET{used\_objects}} & \\
%\quad \texttt{\BLUE{SET} \VIOLET{source\_items} = all items of \VIOLET{source\_object}} & \\
%\quad \texttt{\BLUE{SET} \VIOLET{combs\_items} = all combinations of 2 or more items in \VIOLET{object1}} & \\
\quad \texttt{\BLUE{REPEAT} (for $N_{\rm br\_trials}$ times)  } & \\
\quad \quad \texttt{\BLUE{RAND SET} \VIOLET{comb} as a combination of 2 or more items in \VIOLET{object1}} & \\
%\quad \quad \texttt{\BLUE{SET} \VIOLET{combs\_dest\_objs} = all combinations of |\VIOLET{comb}| objects} & \\
\quad \quad \texttt{\BLUE{REPEAT} (for $N_{\rm br\_trials}$ times) } & \\
\quad \quad \quad \texttt{\BLUE{RAND SET} \VIOLET{dest\_objects} as a combination of $\magn{\mathtt{\VIOLET{comb}}}$ objects in $\mathcal{J}$} & \\
\quad \quad \quad \texttt{\BLUE{FOR} (each item j of \VIOLET{comb})} & \\
\quad \quad \quad \quad \texttt{$X'$ = $X$: move item j from \VIOLET{object1} to \VIOLET{dest\_objs}[j]} &  \\
\end{tabular}
\fbox{\colorbox{lightgray}{\begin{minipage}[fill=gray]{0.96\textwidth}
\begin{tabular}{ll}
\quad \quad \quad \texttt{\textcolor{blue}{IF} ($X'$ is $\mathfrak{C}$-admissible and $g(X')$ < $g(X)$} & \rdelim\}{3}{20pt}[\textit{\scriptsize better reply}] \\
\quad \quad \quad \quad \texttt{\BLUE{SET} $X$ = $X'$ and \BLUE{RETURN} $X$} & \\
\quad \quad \quad \texttt{\BLUE{END IF}} & \\
\end{tabular}
\end{minipage}}}
\begin{tabular}{ll}
\quad \quad \quad \texttt{\BLUE{END FOR}} & \\
\quad \quad \texttt{\BLUE{END REPEAT}} & \\
\quad \texttt{\BLUE{END REPEAT}} & \\
\texttt{\BLUE{END REPEAT}} & \\
\texttt{\BLUE{RETURN} $X$} & \\
\end{tabular}
\end{minipage}
}
}
\caption{\texttt{RemoveObject.better\_reply}}
\label{Tb:RemoveObjectBetterReply}
\end{table}
According to this mode, we select a) a source object, b) a combination of items from the source object, and c) a combination of distinct destination objects equal in number to the selected items. Then, each one of the items moves to one of the destination objects, so that the cost of the optimization is reduced (e.g., see example of Figure~\ref{fig:RemoveObject_operation}).

Similarly, we may define the \texttt{constr\_reply} and \texttt{random\_reply} models of this operations (the same way we did for the \texttt{MoveItem} operation in Section~\ref{sec:MoveItem}).

%------------------------------------------------------
\subsection{\texttt{ReverseRemoveObject} operation}	\label{sec:ReverseRemoveObject}

\begin{figure}[h!]
\centering
\includegraphics[scale=1]{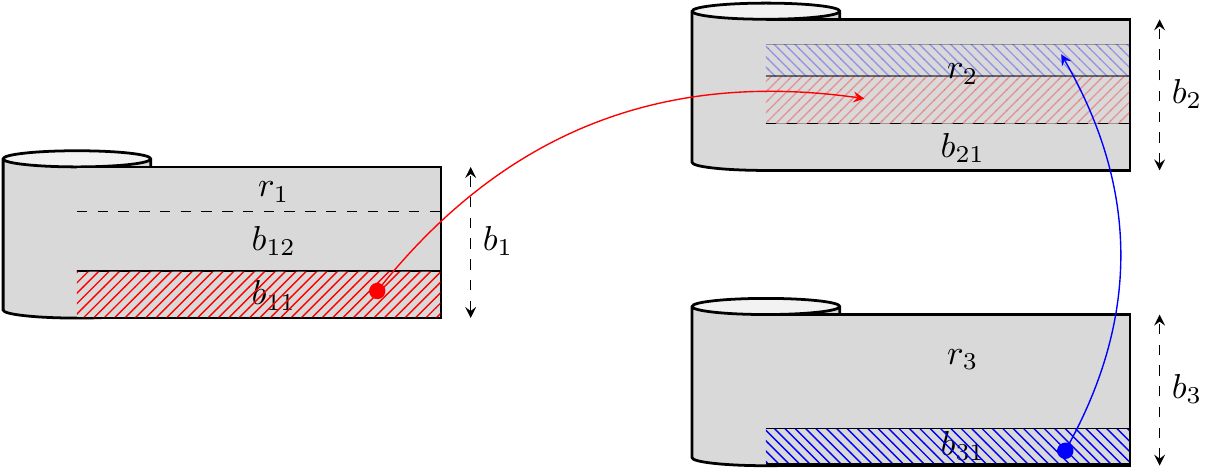}
\caption{\texttt{ReverseRemoveObject} operation.}
\label{fig:ReverseRemoveObject_operation}
\end{figure}

The \texttt{ReverseRemoveObject} operation performs a reverse form of the \texttt{RemoveObject} operation. In particular, the goal of this operation is to combine items from several source objects and move them into a new destination object, see Figure~\ref{fig:ReverseRemoveObject_operation}. % The \texttt{random\_reply()} of this operation simply relies on a random selection of a) the source objects, b) the items selected from these objects, and c) the destination object. Depending on the maximum number of items that may fit on a single object, several types of this operation may be defined. 

The \texttt{better\_reply} mode of this operation is described in Table~\ref{Tb:ReverseRemoveObjectBetterReply}. 
\begin{table} [ht!]
\boxed{{\footnotesize
\begin{minipage}{0.98\textwidth}
\begin{tabular}{ll}
\hline
\textbf{function} \texttt{$X=$ ReverseRemoveObject.better\_reply($X$,$\mathfrak{C}$,$g()$)} & \\\hline
\texttt{\BLUE{SET} \VIOLET{used\_objects} = all objects with non-zero items in $X$} & \\
%\texttt{\BLUE{SET} \VIOLET{comb\_used\_objects} = all combinations of 2 \VIOLET{used\_objects}} & \\
\texttt{\BLUE{REPEAT} (for $N_{\rm br\_trials}$ times) } & \\
\quad \texttt{\BLUE{RAND SET} a combination \VIOLET{comb} of 2 objects in \VIOLET{used\_objects} } & \\
\quad \texttt{\BLUE{REPEAT} (for $N_{\rm br\_trials}$ times) } & \\
\quad \quad \texttt{\BLUE{RAND SET} \VIOLET{dest\_object} from all objects not in \VIOLET{comb} } & \\
\quad \quad \texttt{\BLUE{SET} $X'$ = $X$ } & \\
\quad \quad \texttt{\BLUE{FOR} (each \VIOLET{source\_object} in \VIOLET{comb}) } & \\
\quad \quad \quad \texttt{\BLUE{RAND SET} \VIOLET{item} from the items of \VIOLET{source\_object}} & \\
\quad \quad \quad \texttt{$X'$ = $X'$: move \VIOLET{item} from \VIOLET{source\_object} to \VIOLET{dest\_object}} & \\
\quad \quad \texttt{\BLUE{END FOR}} & \\
\end{tabular}
\fbox{\colorbox{lightgray}{\begin{minipage}[fill=gray]{0.96\textwidth}
\begin{tabular}{lll}
\quad \quad \texttt{\textcolor{blue}{IF} ($X'$ is $\mathfrak{C}$-admissible and $g(X')$ < $g(X)$) } & \rdelim\}{3}{20pt}[\textit{\scriptsize better reply}] \\
\quad \quad \quad \texttt{\BLUE{SET} $X$ = $X'$ and \BLUE{RETURN} $X$} &  & \\
\quad \quad \texttt{\BLUE{END IF}} &  & \\
\end{tabular}
\end{minipage}}}\\
\begin{tabular}{ll}
\quad \texttt{\BLUE{END REPEAT}} & \\
\texttt{\BLUE{END REPEAT}} & \\
\texttt{\BLUE{RETURN} $X$} & \\
\end{tabular}
\end{minipage}
}
}
\caption{\texttt{ReverseRemoveObject.better\_reply}}
\label{Tb:ReverseRemoveObjectBetterReply}
\end{table}
According to this mode, we find a) a combination of source objects, b) an item from each source object, and c) a destination object. We, then, move the selected items to the destination object and we check whether the cost has been reduced. 

The \texttt{constr\_reply} and \texttt{random\_reply} modes of this operation can be defined in a similar way, as we did for the \texttt{MoveItem} operation in Section~\ref{sec:MoveItem}.

%------------------------------------------------------
\subsection{\texttt{RemoveItem} operation}	\label{sec:RemoveItem}

The \texttt{RemoveItem} operation checks whether an item assigned to an object can be removed without violating any of the constraints (i.e., job-admissibility and rest-width constraints). Obviously if any such item can be removed without violating the constraints, then it is likely that the objective function will be reduced. For example, this might be the case when there are objects with a single item assigned to them. The \texttt{better\_reply} mode of this operation is described in Table~\ref{Tb:RemoveItemBetterReply}. 
\begin{table} [ht!]
\boxed{{\footnotesize
\begin{minipage}{0.98\textwidth}
\begin{tabular}{ll}
\hline
\textbf{function} \texttt{$X$ = RemoveItem.better\_reply($X$,$\mathfrak{C}$,$g()$)} & \\\hline
\texttt{\BLUE{SET} \VIOLET{used\_objects} = all objects with non-zero items in $X$} & \\
\texttt{\BLUE{REPEAT} (for $N_{\rm br\_trials}$ times) } & \\
\quad \texttt{\BLUE{RAND SET} \VIOLET{object} from \VIOLET{used\_objects}} & \\
% \quad \texttt{\BLUE{SET} \VIOLET{items} = all items of object} & \\
\quad \texttt{\BLUE{FOR} (each \VIOLET{item} of \VIOLET{object}) } & \\
\quad \quad \texttt{$X'$ = $X$: remove \VIOLET{item} from \VIOLET{object}} & \\
\end{tabular}
\fbox{\colorbox{lightgray}{\begin{minipage}[fill=gray]{0.96\textwidth}
\begin{tabular}{lll}
\quad \texttt{\textcolor{blue}{IF} ($X'$ is $\mathfrak{C}$-admissible and $g(X')$ < $g(X)$) } & \rdelim\}{3}{20pt}[\textit{\scriptsize better reply}] \\
\quad \quad \texttt{\BLUE{SET} $X$ = $X'$ and \BLUE{RETURN} $X$} &  & \\
\quad \texttt{\BLUE{END IF}} &  & \\
\end{tabular}
\end{minipage}}}\\
\begin{tabular}{ll}
\quad \texttt{\BLUE{END FOR}} & \\
\texttt{\BLUE{END REPEAT}} & \\
\texttt{\BLUE{RETURN} $X$} & \\
\end{tabular}
\end{minipage}
}
}
\caption{\texttt{RemoveItem.better\_reply}}
\label{Tb:RemoveItemBetterReply}
\end{table}
Similar is the reasoning in \texttt{constr\_reply} and \texttt{random\_reply} modes of this operation.

%%%%%%%%%%%%%%%%%%%%%%%%%%%%%%%%%%%%%%%%%%%%%%%%%%%%%%%%%%%%%%%%%%%%%%%%%%%%%%%%%%%%%%%
\section{Diversification Strategies}		\label{sec:DiversificationStrategies}

The proposed structure of the optimization algorithm may naturally incorporate several \emph{diversification strategies}. Diversification strategies aim at preventing the search process getting trapped in local optima. In this section, we present a set of possible diversification strategies that can be exploited.

Given that optimization problems of the form (\ref{eq:Optimization}) may differ significantly from each other, both in the number of feasible solutions as well as in the search space, the development of a unified selection and tuning of diversification strategies becomes almost impossible. The goal of this paper is to demonstrate that the diversification strategies considered here can make a difference at least on average.

We may group the set of diversification strategies considered here in the following categories.

\begin{itemize}
\item Diverse initialization strategies
\item Bounded-distance dynamic restarts
\item Random perturbations
\item Parallelization
\item Cost-free operations
% \item Cost-free operations with maximum residual-bands
\end{itemize}

\subsection{Diverse initialization strategies} 	\label{sec:AlternativeInitializationStrategies}

As we discussed in Section~\ref{sec:Initialization}, we introduced three alternative criteria in the formulation of initial candidate solutions. Our objective is to generate initial admissible solutions that a) are close enough to the optimal solution, and b) are diverse enough to increase the possibility for convergence to the optimal solution.

\subsection{Bounded-distance dynamic restarts}		\label{sec:DynamicRestarts}

A candidate solution may currently be located at search spaces with no significant potential in reducing the current cost. Thus, in order to avoid stagnation, we allow such candidate solution to be replaced by another one from an earlier stage but within a bounded fitness distance from the current best. There are two operations that indirectly control the process of dynamic restarts, namely the \texttt{good\_standing} and \texttt{high\_potential} functions. As already described in Section~\ref{sec:Filter}, the \texttt{good\_standing} function and its parameters (i.e., $D_{\rm gs}^*$, $G_{\rm gs}^*$ and $N_{\rm gs}^*$) control the criteria for dropping a candidate solution, while the \texttt{high\_potential} function and its parameters (i.e., $D_{\rm hp}^*$, $G_{\rm hp}^*$ and $N_{\rm hp}^*$) control the criteria for reserving a candidate solution. Apart from these parameters, the size of the main and reserve pool also affect the evolution of the optimization algorithm. These parameters need to be carefully tuned so that the bounded-distance dynamic restarts increase the probability of escaping from a local optimum without though delaying significantly the optimization process. The response to these parameters may differ significantly between application scenarios.

\subsection{Random perturbations}

\emph{Random perturbations} constitute an alternative way to avoid stagnation, through the introduction of a sequence of randomly generated operations. In particular, with a small positive probability $\lambda>0$, a candidate solution goes through the \texttt{perturb\_worker}, in which a sequence of operations is randomly selected, and a random initialization of this operation is set (\texttt{random\_reply} mode). If the perturbation leads to an admissible solution, then the modification is accepted independently of the resulting objective value. This is essentially similar in spirit with the \emph{random-walk extension} introduced in \cite{Hoos05}.

\subsection{Parallelization}		\label{sec:Diversification:Parallelization}

The implementation of the optimization algorithm depicted in Figure~\ref{fig:architecture} suggests a high-level (domain-specific) parallel architecture that does not fit to standard parallel patterns. Simple for-loop parallelization, using, e.g., %OpenMP~\cite{OpenMP} or 
a parallel-for pattern, is not possible, since the iteration space is not known a-priori. In fact, we do not know how many times a candidate solution should pass through the optimization worker (\texttt{local\_opt\_worker}) before we stop processing it. 

\begin{figure}[t!]
\centering
\includegraphics[scale=0.8]{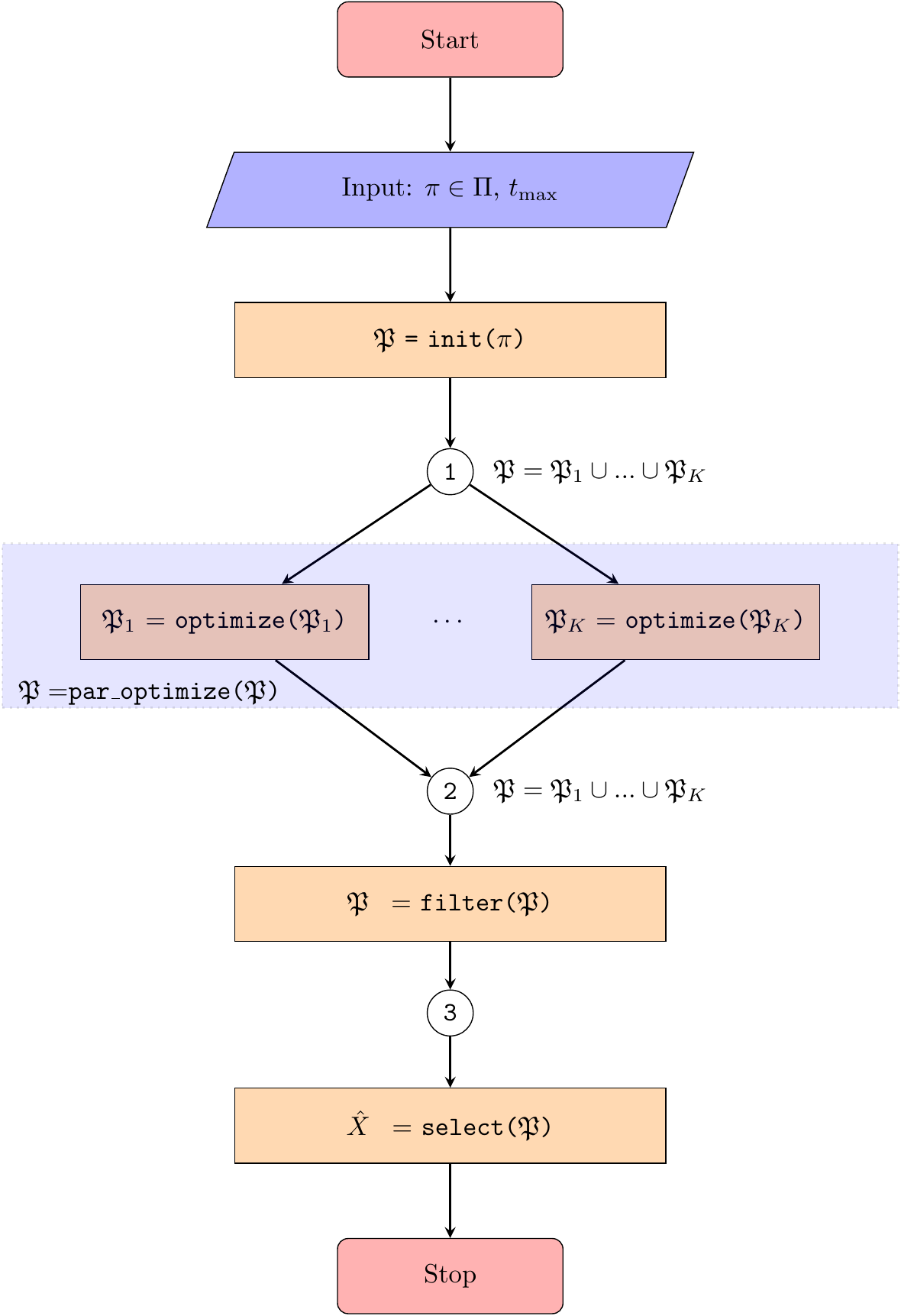}
\caption{Parallel architecture.}
\label{fig:ParallelArchitecture}
\end{figure}

To this end, first we partition the main pool $\mathfrak{P}$ into $K$ sets of candidate solutions (where $K$ is our planned parallel degree). We then process these sets in parallel through the \texttt{optimize} function. Finally, we decide on whether a candidate solution should continue processing through the implementation of the \texttt{filter}. The proposed parallelization architecture is depicted in Figure~\ref{fig:ParallelArchitecture}.

The proposed parallel pattern is a modification of the so-called \emph{pool-pattern}, first implemented in \cite{rosbory_parallelization_2013} and incorporated in the FastFlow parallel programming framework in \cite{aldinucci_pool_2016}. The modification lies in the introduction of the reserve pool $\mathfrak{R}$ that simplifies the implementation of the bounded-distance dynamic restarts, which is an essential part of several SLS algorithms. 

As probably expected, the larger the parallel degree, the larger the probability of getting closer to the optimal solution. This will be demonstrated through experiments in the forthcoming Section~\ref{sec:ExperimentalEvaluation}, but it may also be supported by the statistical literature \cite{Rohatgi76} as pointed out in \cite{Hoos05}.

\subsection{Cost-free operations}

By \emph{cost-free operations} we mean operations that may result in (practically) the same cost. The role of such operations may be proven rather important mainly because they introduce perturbations with no impact to the objective. Thus, through such operations, we may diversify the search of optimal solutions without necessarily restarting from worse assignments. More formally, let $X'$ be a modification over an initial candidate solution $X$. A modification becomes acceptable subject to a better-reply criterion if the following smooth condition is satisfied:
\begin{equation}
g(X') < g(X) + \zeta(X,X'),
\end{equation}
for some real-valued function $\zeta$. Function $\zeta$ applies a form of extra allowance in the acceptance of an improvement. For example, we may assume that $\zeta=\epsilon$ for some small positive constant $\epsilon>0$. Alternative criteria may be defined.

%%%%%%%%%%%%%%%%%%%%%%%%%%%%%%%%%%%%%%%%%%%%%%%%%%%%%%%%%%%%%%%%%%%%%%%%%%%%%%%%%%%%%%%
\section{Experimental Evaluation} 		\label{sec:ExperimentalEvaluation}

The goal in this section is to evaluate the behavior of the evolutionary algorithm and understand the role of the alternative diversification strategies introduced. In particular, we are going to demonstrate the effect of each one of the proposed diversification strategies in the overall performance of the algorithm. To this end, we consider a number of optimization problems of the form (\ref{eq:Optimization}) borrowed from the industry of manufacturing transformer cores (cf.,~\cite{chasparis_optimization_2016}). Each optimization problem is fully described by a set of available rolls and a set of required items to be produced (characterized by their width and weight). The optimization problems considered span from jobs of small number of items  to medium or large number of items, thus covering a range of possibilities and enough variety to test the effect of the proposed algorithm and the diversification strategies.

The optimization problem (\ref{eq:Optimization}) minimizes the used objects' weight. Since the performance of the optimization might be affected by the size of the problem (i.e., the weight of the produced items), we introduce a performance metric normalized by the size of the problem. Let us assume that a set $\Pi$ of different optimization problems is considered. Let $\hat{X}_{\pi}$ denote the best estimate discovered by the algorithm, and $W_{\pi}$ be the total weight of the required items, i.e., $W_{\pi}\df\sum_{i\in\mathcal{I}}w_i$. We introduce the following evaluation metric, $\mathcal{G}:\Pi\to\mathbb{R}_+$, such that 
\begin{equation}	\label{eq:PerformanceMetric}
\mathcal{G}(\pi) \df \frac{g(\hat{X}_{\pi})}{W_{\pi}}\frac{g(\hat{X}_{\pi})}{\sum_{\pi\in\Pi}g(\hat{X}_{\pi})}.
\end{equation}
The first part of the performance metric corresponds to the ratio of objects' weight over the items' weight. Naturally the higher this ratio is, the less efficient the solution would be. However, an inefficient solution at a small-size problem does not have the same impact with an inefficient solution at a large-size problem. To this end, the role of the second ratio is to normalize the effect of the solution in the overall cost.
Furthermore, note that $\sum_{\pi\in\Pi}\mathcal{G}(\pi)$ is the weighted average ratio of the used objects' weight over the required items' weight, i.e., it captures the efficiency of the optimization. 

In the following subsections, we will investigate the effect of the alternative diversification strategies proposed here (as well as some of the optimization parameters) and their impact in the overall performance of the algorithm via metric (\ref{eq:PerformanceMetric}).

%%~~~~~~~~~~~~~~~~~~~~~~~
%\subsection{Random-mode}
%
%The following experiment was performed under Probability Mutations = 0 and Sigmoid Parameter = 1, Gradient = Varying.
%
%\begin{figure}[h!]
%\centering
%\input{./Figures/Effect_RandomMode.tex}
%\caption{The effect of Random Mode.}
%\end{figure}
%
%I guess one of the important conclusions of this graph is that mutations may help reducing the cost, but only when combined with the Random-Mode operations. When Random-Mode operations are not included, then Mutations have no impact.
%
%Note that for higher parallelization degree, the effect of the random mode becomes more evident, as compared to low parallelization degree.

%~~~~~~~~~~~~~~~~~~~~~~~
\subsubsection*{Diverse initialization strategies}

\begin{figure}[h!]
\centering
\includegraphics[scale=1]{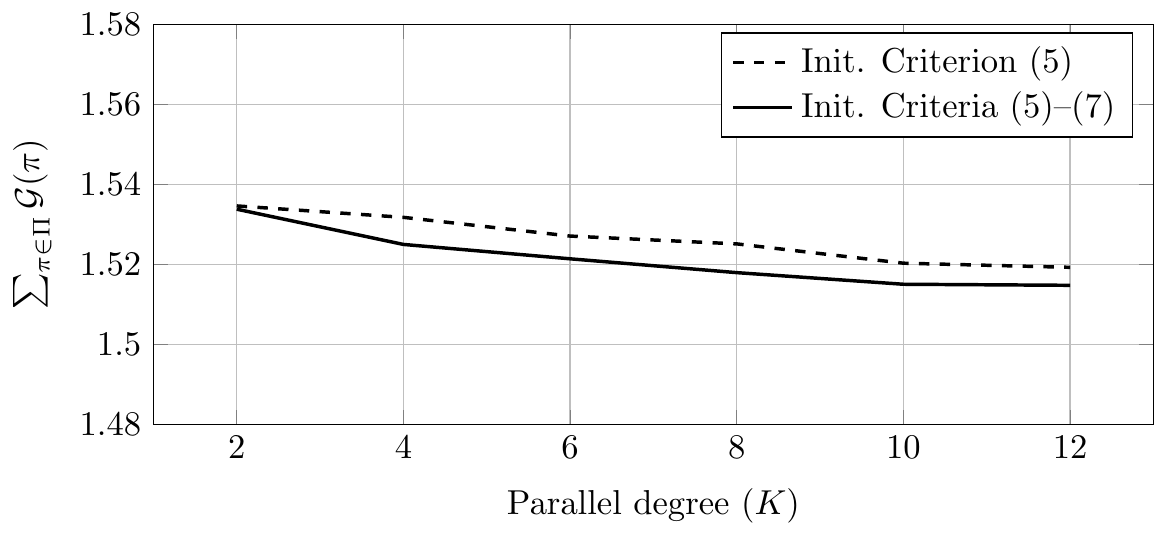}
\caption{The effect of initialization strategies.}
\label{fig:Effect_FFDs}
\end{figure}

We start our investigation by varying the set of strategies considered in the initialization phase of the algorithm. In Figure~\ref{fig:Effect_FFDs}, we see the impact of the initially considered candidate solutions when we restrict the set of strategies used to generate them. In the first scenario (dashed line), we employ the initialization criterion of Equation~(\ref{eq:InitializationFitness1}), while in the second scenario (solid line), we consider all available initialization criteria of Equations~(\ref{eq:InitializationFitness1})--(\ref{eq:InitializationFitness3}). In either case, the initial set of candidate solutions (main pool $\mathfrak{P}$) has the same size and it is populated equally by each one of the considered initialization strategies.

It is important to point out that the benefit of the larger variety of initialization strategies is also due to the fact that the considered strategies employ a form of weight optimization therein. We should not expect that the same improvement occurs when we consider initialization strategies that do not apply a form of weight optimization. 

In Figure~\ref{fig:Effect_FFDs}, as well as in the following experiments, we also demonstrate how the performance varies as we increase the parallel degree ($K$). This corresponds to the effect of parallelization following the architecture of Figure~\ref{fig:ParallelArchitecture}. Furthermore, in all these experiments, $t_{\max}=5min$, unless otherwise specified.

%~~~~~~~~~~~~~~~~~~~~~~~
\subsubsection*{Random perturbations}

\begin{figure}[h!]
\centering
\includegraphics[scale=1]{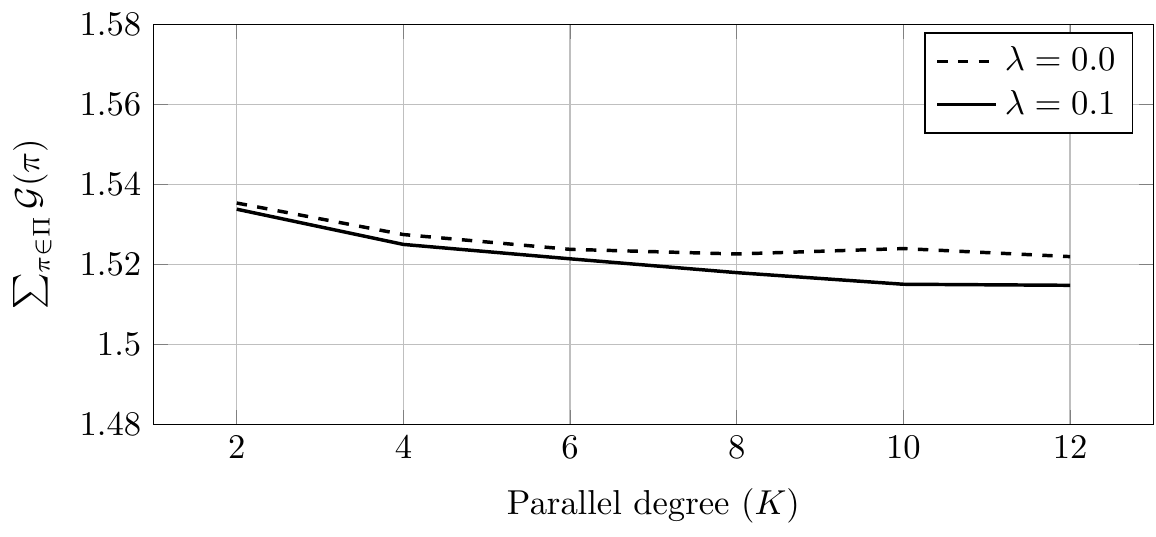}
\caption{The effect of random perturbations.}
\label{fig:Effect_Mutations}
\end{figure}

In this set of experiments, we would like to test the effect of the random perturbations controlled by the parameter $\lambda>0$. Figure~\ref{fig:Effect_Mutations} shows the increase in performance when $\lambda$ increases to $\lambda=0.1$ (i.e., with probability $0.1$, a candidate solution goes through the \texttt{perturb\_worker}. % Within this worker, a sequence of  (randomly initialized) operations are executed that admit the constraints. 
Note that the random perturbations have a positive impact in the optimization performance.

%~~~~~~~~~~~~~~~~~~~~~~~
\subsubsection*{Optimization time}

We would also like to investigate how the optimization algorithm responds when the maximum allowed optimization time ($t_{\max}$) increases. It is natural to expect that the performance in this case should increase, since some of the candidate solutions will be processed for a longer period of time. Note further that the increase in performance seems to reduce as we increase the parallel degree. One explanation for this behavior is the fact that an increased parallel degree already allows for a faster processing, thus it is highly likely that we are already sufficiently close to the optimal solution.

\begin{figure}[h!]
\centering
\includegraphics[scale=1]{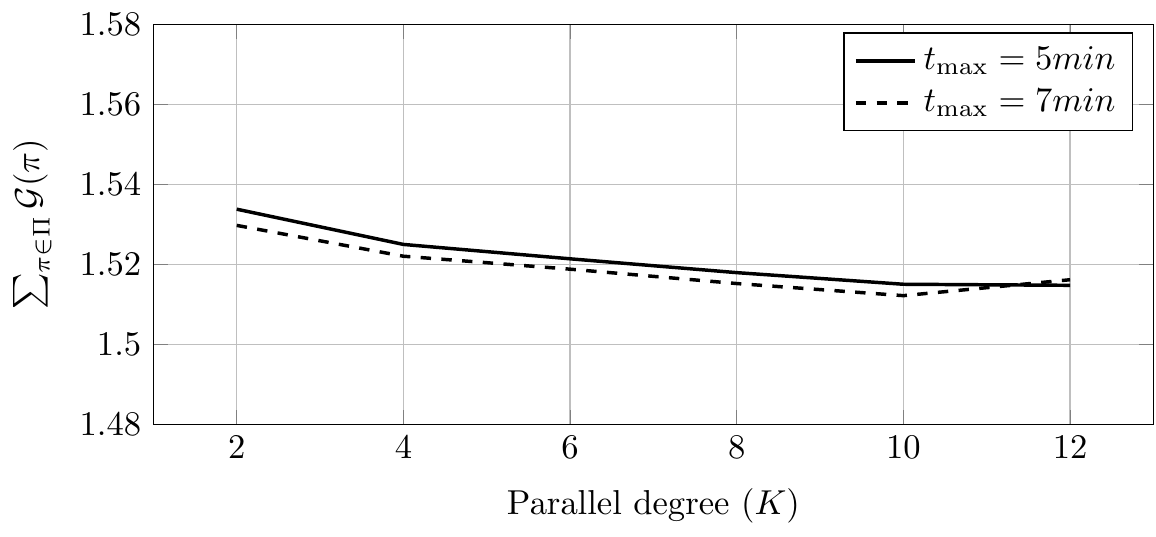}
\caption{The effect of simulation time.}
\end{figure}

%~~~~~~~~~~~~~~~~~~~~~~~
\subsubsection*{High-potential threshold}

Recall that $D_{\rm hp}^*$ and $G_{\rm hp}^*$ are two thresholds that control the reservation of candidate solutions into the reserve pool $\mathfrak{R}$. We would like here to investigate the effect of $D_{\rm hp}^*$ which corresponds to the normalized fitness-distance from the current best within which a high-potential candidate solution can be stored into the reserve pool $\mathfrak{R}$. In other words, we would like to investigate how the performance of the optimization varies when we move the reservation threshold towards earlier stages.

\begin{figure}[h!]
\centering
\includegraphics[scale=1]{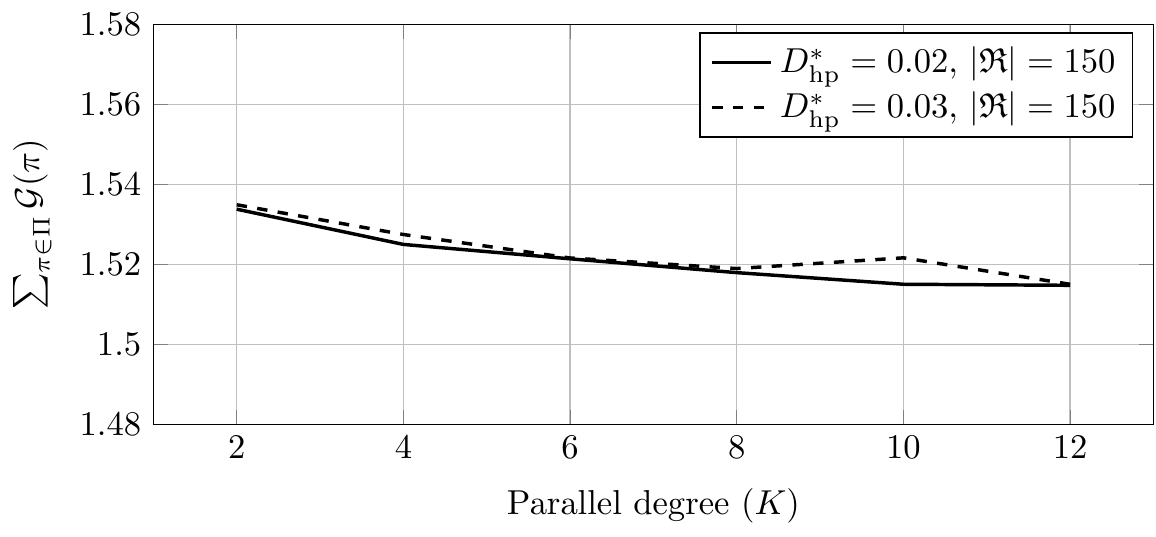}
\caption{The effect of high-potential threshold.}
\label{fig:Effect_AddReservation}
\end{figure}

\begin{figure}[h!]
\centering
\includegraphics[scale=1]{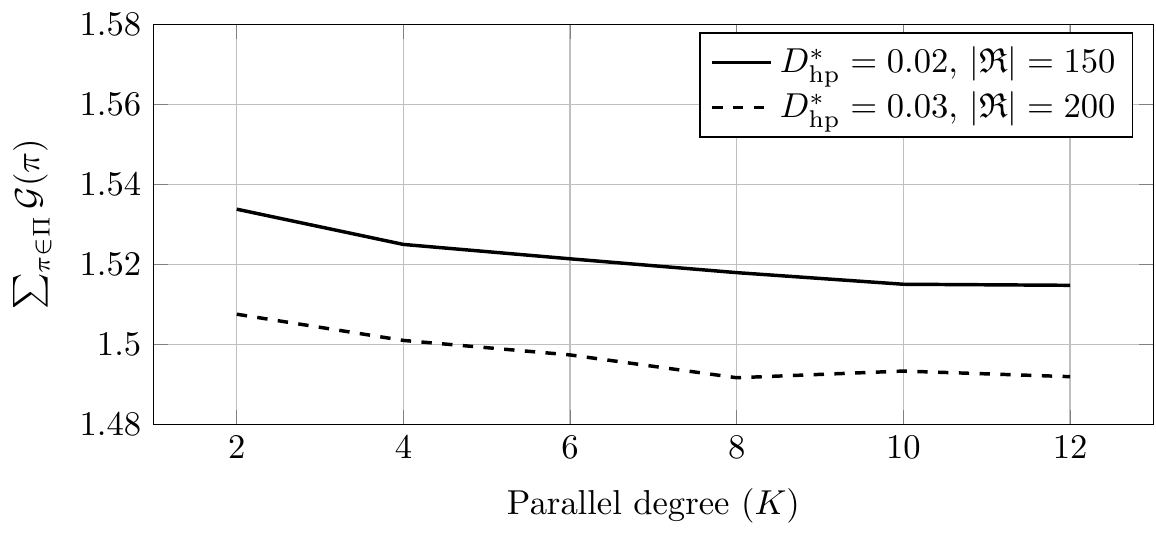}
\caption{The effect of high-potential threshold as we increase the size of the reserve pool.}
\label{fig:Effect_AddReservationWReservePool}
\end{figure}

When candidate solutions from earlier stages are stored into the reserve pool $\mathfrak{R}$, we should expect that the escape probability from local optima should increase. However, the size of the reserve pool $\mathfrak{R}$ is fixed, which means that the oldest candidate solutions may be dropped to create space for more recent reservations. Thus, when we just increase the threshold $D_{\rm hp}^*$ the benefit is not guaranteed (Figure~\ref{fig:Effect_AddReservation}). However, if we also increase the size of the reserve pool, then the benefit for increasing the threshold can be materialized (Figure~\ref{fig:Effect_AddReservationWReservePool}).

%~~~~~~~~~~~~~~~~~~~~~~~
\subsubsection*{Good-standing threshold}

Recall that $D_{\rm gs}^*$ and $G_{\rm gs}^*$ are two thresholds that control the assessment over allowing a candidate solution to continue processing (Section~\ref{sec:Filter}). In particular, $D_{\rm gs}^*$ denotes the normalized fitness distance from the current best within which a candidate solution is allowed to continue processing. We wish to investigate how the performance of the optimization will change as we increase $D_{\rm gs}^*$, i.e., if we allow candidate solutions to remain longer within the main pool $\mathfrak{P}$. 

\begin{figure}[h!]
\centering
\includegraphics[scale=1]{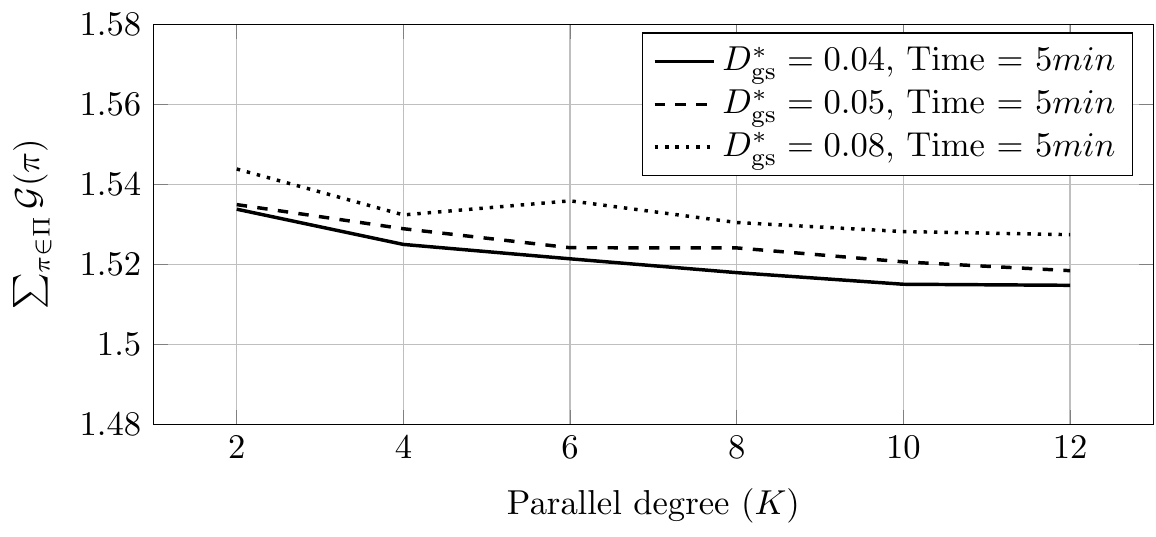}
\caption{The effect of good-standing threshold.}
\label{fig:Effect_KeepProcessing}
\end{figure}

\begin{figure}[h!]
\centering
\includegraphics[scale=1]{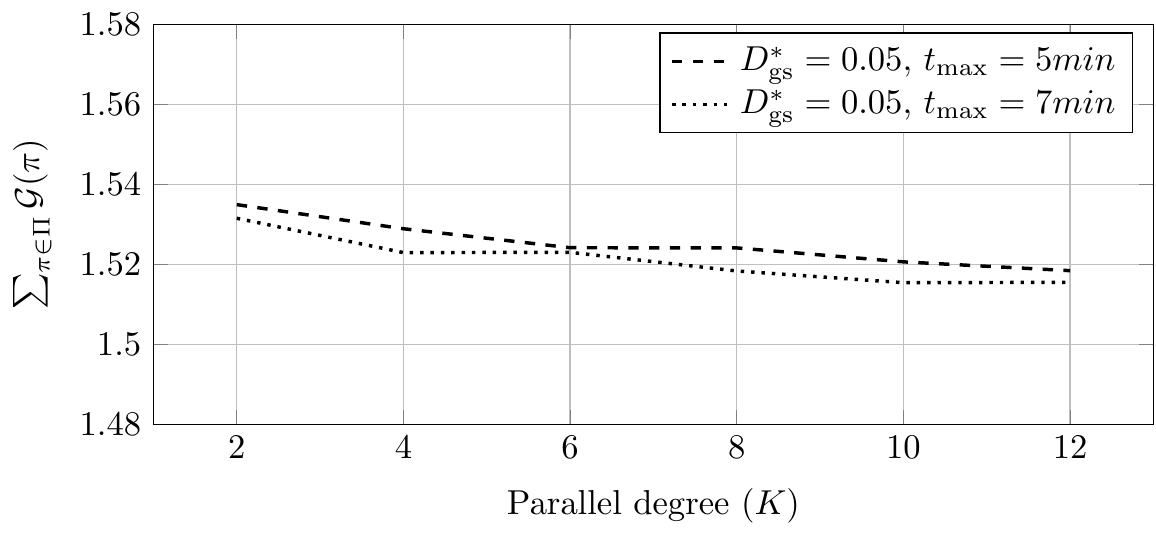}
\caption{The effect of the good-standing threshold.}
\label{fig:Effect_KeepProcessingWTime}
\end{figure}

Figure~\ref{fig:Effect_KeepProcessing} demonstrates how the performance changes as we progressively increase $D_{\rm gs}^*$. As expected, the longer we allow a candidate solution to remain within the main pool, the slower the rate of using reserved candidate solutions, and the greater the probability that the process is trapped in a local optimum. Thus, we should expect that the performance may degrade as we increase $D_{\rm gs}^*$ more than necessary (Figure~\ref{fig:Effect_KeepProcessing}). However, if we also increase the optimization time, then we may achieve better performance (since we exploit better the fact that solutions remain in the pool for longer periods of time), Figure~\ref{fig:Effect_KeepProcessingWTime}.

%~~~~~~~~~~~~~~~~~~~~~~~
\subsubsection*{Cost-free operations}

\begin{figure}[h!]
\centering
\includegraphics[scale=1]{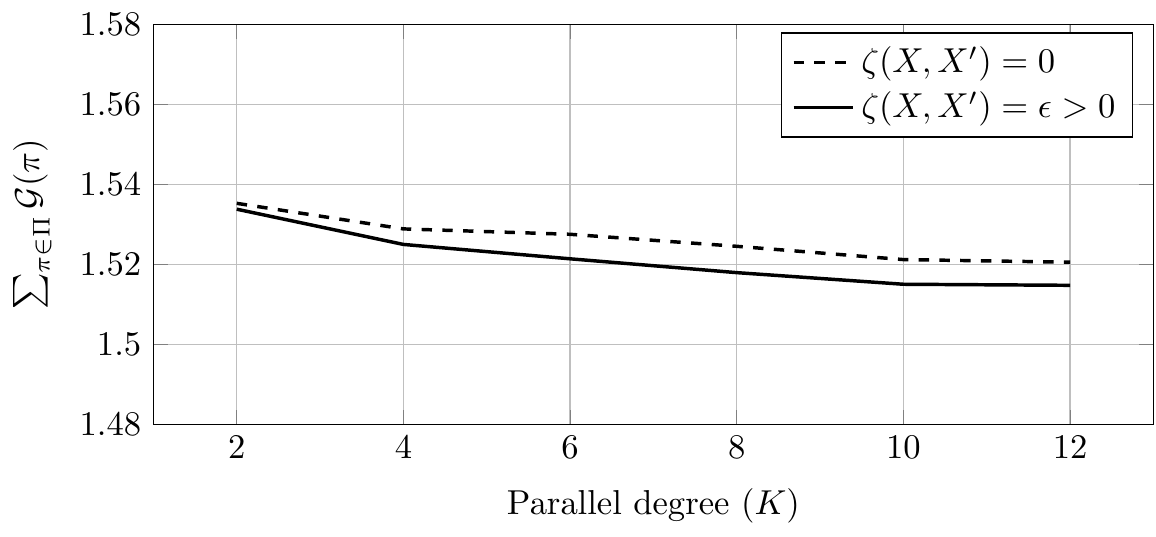}
\caption{The effect of cost-free operations.}
\label{fig:CostFreeOperations}
\end{figure}

In the last experiment, we wish to investigate the effect of cost-free operations in the performance of the optimization. Cost-free operations essentially allow for local operations that do not have any significant impact in the overall performance. As probably expected, this allows for a candidate solution to escape from a local optimum with no impact in the performance. Figure~\ref{fig:CostFreeOperations} demonstrates the fact that cost-free operations can indeed increase the overall performance rather constantly over the parallel degree.

%%~~~~~~~~~~~~~~~~~~~~~~~
%\subsection{Constant vs Varying Gradient}
%
%This experiment was performed under the following conditions Probability Random Mode = 0.2, Probability of Mutations = 0.1, Sigmoid Parameter = 1.
%
%Maybe we should also compare it under the conditions of Probability Random Mode = 0.0 and Probability Mutations = 0.0.
%
%\begin{figure}[h!]
%\centering
%\input{./Figures/Effect_VaryingGradient.tex}
%\caption{The effect of Maximum Rest-Band Criterion.}
%\end{figure}

%%~~~~~~~~~~~~~~~~~~~~~~~
%\subsection{Cost-free operations with maximum residual bands}
%
%This experiment was executed when Prob Random Mode = 0.2, Prob Mutations = 0.1, Sigmoid Parameter = 1, Gradient = Varying.
%
%!!! Maybe we also need to run it for the case of Prob Random Mode = 0.0, Prob Mutations = 0.0, Sigmoid Parameter = 1, Gradient = Varying.
%
%\begin{figure}[h!]
%\centering
%\input{./Figures/Effect_MaximumRestBandCriterion.tex}
%\caption{The effect of Maximum Rest-Band Criterion.}
%\end{figure}

%%%%%%%%%%%%%%%%%%%%%%%%%%%%%%%%%%%%%%%%%%%%%%%%%%%%%%%%%%%%%%%%%%%%%%%%%%%%%%%%%%%%%%%%%%%%%%%%%%%
\section{Conclusions}		\label{sec:Conclusions}

This paper presented an evolutionary SLS algorithm specifically tailored for one-dimensional cutting-stock optimization problems. The novelty of the proposed algorithm lies in its architecture that allowed for the introduction of a novel parallel pattern and the easy integration of a large family of diversification strategies. Although the proposed SLS algorithm was presented within the context of one-dimensional cutting-stock problems, its architecture is generic enough to be integrated into a larger family of combinatorial optimization problems. Finally, the presented experimental evaluation demonstrated a significant improvement in efficiency due to both the proposed parallel pattern as well as the additional diversification strategies presented.

%%%%%%%%%%%%%%%%%%%%%%%%%%%%%%%%%%%%%%%%%%%%%%%%%%%%%%%%%%%%%%%%%%%%%%%%%%%%%%%%%%%%%%%%%%%%%%%%%%%
%\begin{acknowledgements}
%
%\end{acknowledgements}

% BibTeX users please use one of
%\bibliographystyle{spbasic}      % basic style, author-year citations
%\bibliographystyle{spmpsci}      % mathematics and physical sciences
\bibliographystyle{splncs03}
\bibliography{2017_Heuristics_CoilSlitting_Bibliography.bib}   % name your BibTeX data base

\begin{thebibliography}{10}
\providecommand{\url}[1]{\texttt{#1}}
\providecommand{\urlprefix}{URL }

\bibitem{Aktin09}
Aktin, T., {\"{O}}zdemir, R.G.: {An integrated approach to the one-dimensional
  cutting stock problem in coronary stent manufacturing}. European Journal of
  Operational Research  196(2),  737--743 (2009)

\bibitem{aldinucci_pool_2016}
Aldinucci, M., Campa, S., Danelutto, M., Kilpatrick, P., Torquati, M.: Pool
  {Evolution}: {A} {Parallel} {Pattern} for {Evolutionary} and {Symbolic}
  {Computing}. International Journal of Parallel Programming  44(3),  531--551
  (Jun 2016)

\bibitem{Belov02}
Belov, G., Scheithauer, G.: {A cutting plane algorithm for the one-dimensional
  cutting stock problem with multiple stock lengths}. European Journal of
  Operational Research  141,  274--294 (2002)

\bibitem{Belov06}
Belov, G., Scheithauer, G.: {A branch-and-cut-and-price algorithm for
  one-dimensional stock cutting and two-dimensional two-stage cutting}.
  European Journal of Operational Research  171,  85--106 (2006)

\bibitem{Beraldi09}
Beraldi, P., Bruni, M.E., Conforti, D.: {The stochastic trim-loss problem}.
  European Journal of Operational Research  197(1),  42--49 (2009)

\bibitem{Brooks87}
Brooks, R.L., Smith, C.A.B., Stone, A.H., Tutte, W.T.: The Dissection of
  Rectangles Into Squares, pp. 88--116. Birkh{\"a}user Boston, Boston, MA
  (1987)

\bibitem{Carvalho02}
de~Carvalho, J.V.: {LP} models for bin packing and cutting stock problems.
  European Journal of Operational Research  141,  253--273 (2002)

\bibitem{chasparis_optimization_2016}
Chasparis, G., Zellinger, W., Haunschmid, V., Riedenbauer, M., Stumptner, R.:
  On the optimization of material usage in power transformer manufacturing. In:
  2016 {IEEE} 8th {International} {Conference} on {Intelligent} {Systems}
  ({IS}). pp. 680--685 (Sep 2016)

\bibitem{Cherri14}
Cherri, A.C., Arenales, M.N., Yanasse, H.H., Poldi, K.C., Gon{\c
  c}alves~Vianna, A.C.: The one-dimensional cutting stock problem with usable
  leftovers - {A} survey. European Journal of Operational Research  236(2),
  395--402 (Jul 2014)

\bibitem{Dyckhoff90}
Dyckhoff, H.: A typology of cutting and packing problems. European Journal of
  Operational Research  44,  145--159 (1990)

\bibitem{Gerstl97}
Gerstl, A., Karisch, S.E.: Cost optimization for the slitting of core
  laminations for power transformers. Annals of Operations Research  69,
  157--169 (1997)

\bibitem{Gilmore61}
Gilmore, P., Gomory, R.: A linear programming approach to the cutting stock
  problem. Operations Research  9,  848--859 (1961)

\bibitem{Gilmore63}
Gilmore, P., Gomory, R.: A linear programming approach to the cutting stock
  problem, {P}art {II}. Operations Research  11,  863--888 (1963)

\bibitem{Gradisar99}
Gradi{\v{s}}ar, M., Kljaji{\'{c}}, M., Resinovi{\'{c}}, G., Jesenko, J.: {A
  sequential heuristic procedure for one-dimensional cutting}. European Journal
  of Operational Research  114,  557--568 (1999)

\bibitem{Gradisar05}
Gradi{\v{s}}ar, M., Trkman, P.: {A combined approach to the solution to the
  general one-dimensional cutting stock problem}. Computers and Operations
  Research  32,  1793--1807 (2005)

\bibitem{Haessler91}
Haessler, R.W., Sweeney, P.E.: Cutting stock problems and solution procedures.
  European Journal of Operational Research  54,  141--150 (1991)

\bibitem{Haouari09}
Haouari, M., Serairi, M.: {Heuristics for the variable sized bin-packing
  problem}. Computers and Operations Research  36 (2009)

\bibitem{Hinterding93}
Hinterding, R., Khan, L.: Genetic algorithms for cutting stock problems: With
  and without contiguity. In: Selected Papers from the AI'93 and AI'94
  Workshops on Evolutionary Computation, Process in Evolutionary Computation.
  pp. 166--186. AI \'93/AI \'94, Springer-Verlag, London, UK, UK (1995)

\bibitem{Holthaus02}
Holthaus, O.: {Decomposition approaches for solving the integer one-dimensional
  cutting stock problem with different types of standard lengths}. European
  Journal of Operational Research  141,  295--312 (2002)

\bibitem{Hoos05}
Hoos, H., St{\"{u}}tzle, T.: Stochastic Local Search: Foundations and
  Applications. Elsevier Inc. (2005)

\bibitem{Kallrath14}
Kallrath, J., Rebennack, S., Kallrath, J., Kusche, R.: {Solving real-world
  cutting stock-problems in the paper industry: Mathematical approaches ,
  experience and challenges}. European Journal of Operational Research  238(1),
   374--389 (2014)

\bibitem{Kantorovich60}
Kantorovich, L.: Mathematical methods of organizing and planning production.
  Management Science  6,  366--422 (1960)

\bibitem{Lodi02}
Lodi, A., Martello, S., Monaci, M.: {Two-dimensional packing problems: A
  survey}. European Journal of Operational Research  141(2),  241--252 (2002)

\bibitem{Lu15}
Lu, H.c., Huang, Y.h.: {An efficient genetic algorithm with a corner space
  algorithm for a cutting stock problem in the TFT-LCD industry}. European
  Journal of Operational Research  246(1),  51--65 (2015)

\bibitem{Onwubolu03}
Onwubolu, G.C., Mutingi, M.: {A genetic algorithm approach for the cutting
  stock problem}. Journal of Intelligent Manufacturing  14,  209--218 (2003)

\bibitem{Pentico08}
Pentico, D.W.: {The assortment problem : A survey}. European Journal of
  Operational Research  190,  295--309 (2008)

\bibitem{Peter96}
P{\'{e}}ter, A., Andr{\'{a}}s, A., Zsuzsa, S.: {A Genetic Solution for the
  Cutting Stock Problem} (1996)

\bibitem{Rohatgi76}
Rohatgi, V.: An Introduction to Probability Theory and Mathematical Statistics.
  John Wiley \& Sons, New York, NY (1976)

\bibitem{Romanowski12}
Romanowski, A., Nowotniak, R., Kawecki, K., Jaworski, T., Chaniecki, Z.,
  Grudzie{\'{n}}, K.: {Evolutionary Algorithms Approach for Cutting Stock
  Problem}. Image Processing {\&} Communications  17(4),  297--306 (2012)

\bibitem{rossbory_parallelization_2016}
Rossbory, M., Chasparis, G.: Parallelization of stochastic-local-search
  algorithms using high-level parallel patterns. In: High-Level Programming for
  Heterogeneous and Hierarchical Parallel Systems. Prague, Czech Republic (Jan
  2016)

\bibitem{rosbory_parallelization_2013}
Rossbory, M., Reisner, W.: Parallelization of {Algorithms} for {Linear}
  {Discrete} {Optimization} {Using} {ParaPhrase}. In: 2013 24th {International}
  {Workshop} on {Database} and {Expert} {Systems} {Applications}. pp. 241--245
  (Aug 2013)

\bibitem{Shahin04}
Shahin, A.a., Salem, O.M.: {Using genetic algorithms in solving the
  one-dimensional cutting stock problem in the construction industry}. Canadian
  Journal of Civil Engineering  31,  321--332 (2004)

\bibitem{Shao09}
Shao, X., Li, X., Gao, L., Zhang, C.: {Integration of process planning and
  scheduling - A modified genetic algorithm-based approach}. Computers and
  Operation Research  36,  2082--2096 (2009)

\bibitem{Sweeney90}
Sweeney, P.E., Haessler, R.W.: {One-dimensional cutting stock decisions for
  rolls with multiple quality grades}. European Journal of Operational Research
   44(June 1988),  224--231 (1990)

\bibitem{Umetani03}
Umetani, S., Yagiura, M., Ibaraki, T.: {One-dimensional cutting stock problem
  to minimize the number of different patterns}. European Journal of
  Operational Research  146,  388--402 (2003)

\bibitem{Wascher07}
W{\"{a}}scher, G., Hau{\ss}ner, H., Schumann, H.: {An improved typology of
  cutting and packing problems}. European Journal of Operational Research  183,
   1109--1130 (2007)

\bibitem{williamson_design_2011}
Williamson, D.P., Shmoys, D.B.: The {Design} of {Approximation} {Algorithms}.
  Cambridge University Press, New York, NY, USA, 1st edn. (2011)

\bibitem{Yanasse06}
Yanasse, H.H., Limeira, M.S.: {A hybrid heuristic to reduce the number of
  different patterns in cutting stock problems}. Computers and Operations
  Research  33,  2744--2756 (2006)

\end{thebibliography}

% Non-BibTeX users please use
%\begin{thebibliography}{}
%%
%% and use \bibitem to create references. Consult the Instructions
%% for authors for reference list style.
%%
%\bibitem{RefJ}
%% Format for Journal Reference
%Author, Article title, Journal, Volume, page numbers (year)
%% Format for books
%\bibitem{RefB}
%Author, Book title, page numbers. Publisher, place (year)
%% etc
%\end{thebibliography}

\end{document}
% end of file template.tex